%% file: Halvani-A_Step_Towards_Interpretable_Authorship_Verification.tex
\documentclass{article}
\usepackage[a4paper, total={16cm, 26cm}]{geometry}
\usepackage[linesnumbered,vlined,ruled]{algorithm2e}
\usepackage[parfill]{parskip}
\usepackage[table]{xcolor}
\usepackage[utf8]{inputenc}
\usepackage{adjustbox}
\usepackage{amsmath}
\usepackage{amssymb}
\usepackage{bm,array}
\usepackage{booktabs}
\usepackage{colortbl}
\usepackage{enumerate}
\usepackage{enumitem}
\usepackage{float}
\usepackage{graphicx,dblfloatfix}
\usepackage{hyperref}
\usepackage{cleveref}
\usepackage{makecell}
\usepackage{multirow}
\usepackage{nicefrac}
\usepackage{tabularx}
\usepackage{tikz}
\usepackage{url}
\usepackage{xspace}
\usetikzlibrary{positioning}

\hypersetup
{
	pdftitle={Towards Interpretable Authorship Verification},
	pdfauthor={Oren Halvani, Lukas Graner and Roey Regev},
	pdfsubject={Authorship Verification},
	pdfcreator={},
	pdfproducer={},
	pdfkeywords={Authorship Analysis} {Bias Mitigation} {Interpretation},
	pdfpagemode=UseNone,
	pdfstartview={XYZ null null null},
	colorlinks = true,
	linkcolor  = blue,
	citecolor  = blue,
	urlcolor  = blue,
	anchorcolor = blue
}

\SetCommentSty{tcccolored}

\SetNlSty{mynlfont}{}{}

\newcommand*{\A}            {\mathcal{A}}

\newcommand*{\unknown}      {\mathcal{U}}

\newcommand*{\D}            {\mathcal{D}}
\newcommand*{\DA}           {{\D_{\A}}}

\newcommand*{\Dunk}         {\D_{\,\unknown}}

\newcommand*{\Dset}         {\mathbb{D}}

\newcommand*{\numberOfExistingBaselines}   {ten\xspace}
\newcommand*{\numberOfTestCorpora}         {four\xspace}

\newcommand*{\Corpus}       {\mathcal{C}}

\newcommand*{\CorpusStackEx}   {\Corpus_\mathrm{Stack}}

\newcommand*{\CorpusYelp}      {\Corpus_\mathrm{Yelp}}
\newcommand*{\CorpusReddit}    {\Corpus_\mathrm{Reddit}}

\newcommand*{\CorpusAmazon}    {\Corpus_\mathrm{Amazon}}

\newcommand*{\Model}        {\mathcal{M}}
\newcommand*{\Arefset}      {\Dset_{\A}}

\newcommand*{\Problem}      {c}

\newcommand*{\Threshold}      {\theta}

\newcommand*{\Fset}        {\mathbb{F}}

\newcommand*{\taveer}        {\textsf{TAVeer}\mbox{}\xspace}
\newcommand*{\coav}          {\textsf{COAV}\mbox{}\xspace}
\newcommand*{\occav}         {\textsf{OCCAV}\mbox{}\xspace}

\newcommand*{\hossamAV}      {\textsf{DynamicAV}\mbox{}\xspace}

\newcommand*{\khonjiGI}      {\textsf{ASGALF}\mbox{}\xspace}
\newcommand*{\koppelGI}      {\textsf{GenIM}\mbox{}\xspace}
\newcommand*{\koppelIM}      {\textsf{IM}\mbox{}\xspace}
\newcommand*{\koppelUnmask}  {\textsf{Unmasking}\mbox{}\xspace}
\newcommand*{\genUnmask}     {\textsf{GenUnmasking}\mbox{}\xspace}
\newcommand*{\nealAVIF}      {\textsf{AVIF}\mbox{}\xspace}

\newcommand*{\spatium}       {\textsf{SPATIUM}\mbox{}\xspace}
\newcommand*{\baff}          {\textsf{BAFF}\mbox{}\xspace}
\newcommand*{\stamatatosProf}{\textsf{ProfAV}\mbox{}\xspace}

\newcommand*{\veenmanNNCD}   {\textsf{NNCD}\mbox{}\xspace}
\newcommand*{\AV}       {AV\mbox{}\xspace}

\newcommand*{\av}       {authorship verification\mbox{}\xspace}

\newcommand*{\listStylePatterns}   {\mathcal{L}_{\textrm{TA}}}

\newcommand*{\classY}      {\texttt{Y}\mbox{}\xspace} 
\newcommand*{\classN}      {\texttt{N}\mbox{}\xspace} 
\newcommand*{\unanswered}  {\texttt{U}\mbox{}\xspace}

\newcommand*{\fOne}        {F$_{1}$\mbox{}\xspace}

\newcommand*{\auc}         {AUC\mbox{}\xspace}

\newcommand*{\dmax} {d_{\e{max}}}

\newcommand*{\stateOfTheArt} {state of the art\mbox{}\xspace}

\newcommand*{\posTags}       {POS tags\mbox{}\xspace}
\newcommand*{\charNgrams}    {character $n$-grams\mbox{}\xspace}

\renewcommand  {\quote}[1]{{``#1''}}

\newcommand{\e}[1]{\emph{#1}} 
\newcommand*{\eg}            {e.\,g.,\mbox{}\xspace}
\newcommand*{\ie}            {i.\,e.,\mbox{}\xspace}

\newcommand*{\listY}        {$\mathcal{L}_{\,\classY}$\mbox{}\xspace}
\newcommand*{\listN}        {$\mathcal{L}_{\,\classN}$\mbox{}\xspace}

\newcommand*{\featureImp}{\ensuremath{\lambda}\xspace}

\definecolor{Black}{HTML}{000000}
\definecolor{Ycolor}{HTML}{1a9641}
\definecolor{Ncolor}{HTML}{d7191c}

\definecolor{LightGray}{gray}{0.85}

\makeatletter
\newcommand*{\rom}[1]{\expandafter\@slowromancap\romannumeral #1@}
\makeatother

\begin{document}
\title{\textbf{A Step Towards Interpretable\\ Authorship Verification}}

\author{\textbf{Oren Halvani\footnote{Corresponding author.}, Lukas Graner and Roey Regev} \\ 
	Fraunhofer Institute for Secure Information Technology SIT,\\ Rheinstr. 75, 64295 Darmstadt, Germany\\
	\{\texttt{FirstName.LastName\}@SIT.Fraunhofer.de}}

\date{\vspace{-2ex}}
\maketitle

\begin{abstract}
A central problem that has been researched for many years in the field of digital text forensics is the question whether two documents were written by the same author. Authorship verification (AV) is a research branch in this field that deals with this question. Over the years, research activities in the context of AV have steadily increased, which has led to a variety of approaches trying to solve this problem. Many of these approaches, however, make use of features that are related to or influenced by the topic of the documents. Therefore, it may accidentally happen that their verification results are based not on the writing style (the actual focus of AV), but on the topic of the documents. To address this problem, we propose an alternative AV approach that considers only topic-agnostic features in its classification decision. In addition, we present a post-hoc interpretation method that allows to understand which particular features have contributed to the prediction of the proposed AV method. To evaluate the performance of our AV method, we compared it with ten competing baselines (including the current state of the art) on four challenging data sets. The results show that our approach outperforms all baselines in two cases (with a maximum accuracy of 84\%), while in the other two cases it performs close to the strongest baseline.
\end{abstract}
%====================================================================================================
% Sections
%====================================================================================================
\input{texfiles/Introduction}
\input{texfiles/PreviousWork}
\input{texfiles/ProposedApproach}
\input{texfiles/ProposedInterpretationScheme}
\input{texfiles/Experiments}
\input{texfiles/Conclusions}

\input{texfiles/Acknowledgments}
%====================================================================================================
\bibliographystyle{plain}
\bibliography{Bibliography}
\end{document}

%% file: texfiles/Introduction.tex
\section{Introduction} \label{Introduction}
With the constant increase of documents worldwide, more and more possibilities of identity misuse are becoming established. One example of such identity abuse is 
\quote{CEO Fraud} -- a sophisticated email scam -- in which an attacker sends an email to an employee on behalf of a CEO to perform a specific action (\eg transferring money or sending confidential company information). Another form of identity abuse occurs in the context of compromised accounts, where the attacker distributes messages in the name of the victim. In addition, identity abuse can occur in fake reviews in which, for example, an attempt is made on behalf of an alleged person to positively advertise a product or service provider. One countermeasure that can be applied in all these scenarios is to compare the writing style of the questioned documents to the writing style of those documents, of which the true author $\A$ is known. By this, the question can be answered (with a certain degree of probability) whether the unknown document was also written by $\A$. The comparison of documents based on their writing style is particularly relevant if no other metadata are available to clarify the identity of the unknown author. 

Authorship verification (\AV) -- a branch of research in digital text forensics -- has been dealing with this question for over two decades. 
Technically, \AV represents a similarity detection problem, where for an unknown document $\Dunk$ and a known document $\DA$ it has to be determined whether both were written by the same author $\A$. Here, the focus of the similarity determination lies on the \textbf{writing style} of the documents and not on other factors such as the \textbf{topic} or \textbf{genre}. Otherwise, if $\Dunk$ and $\DA$ have the same topic but were written by two different authors, an AV method would clearly miss its intended goal. A large number of existing AV methods including   \cite{BrocardoStylometryAV:2013,CastroAVAverageSimilarity:2015,KoppelWinter2DocsBy1:2014,NealAVviaIsolationForests:2018,StamatatosPothaImprovedIM:2017,PothaStamatatosExtrinsicAV:2019} make use of \charNgrams (overlapping character sequences), which are known to be closely associated to particular content words and, therefore, can be problematic when dealing with authorship \cite{KoppelComputationalMethodsAA:2009}. Style analysis, however, must abstract from content and focus on content-independent formal properties of linguistic expressions in a text \cite{GamonAuthorshipClassification:2004}. In the light of this conclusion, we propose an alternative approach which, by design, considers only such text units that reflect valid stylistic markers. Our contribution in this paper is threefold: First, we propose a number of topic-agnostic feature categories that effectively quantify the writing style of documents. Second, we propose a transparent \AV method that can be applied to challenging \AV tasks. These include cases, where $\Dunk$ and $\DA$ consist of only a few sentences or cases, in which both documents differ thematically. Third, we propose a post-hoc interpretation method, which allows to understand which features contributed to the prediction of our \AV method. 

The remainder of the paper is organized as follows. Section~\ref{ExistingApproaches} discusses previous work in the context of \AV. In Section~\ref{Proposed_FeatureCategories}, we propose a number of feature categories, which will be used by our \AV method introduced in Section~\ref{ProposedApproach}. In Section~\ref{TAVeer_InterpretationScheme}, we then describe our post-hoc interpretation method. Afterwards, we present our experimental evaluation in Section~\ref{Evaluation} and, finally, in Section~\ref{Conclusions} we conclude the work and provide ideas for future work.

%% file: texfiles/PreviousWork.tex
\section{Previous Work} \label{ExistingApproaches} 
The core of every \AV method is a classification model that aims to decide whether a questioned document $\Dunk$ was written by a certain author $\A$, for which a set $\Arefset = \{ \D_1, \D_2, \ldots \}$ of reference documents is given. With regard to their classification models, we have identified three categories of \AV methods in our previous research work \cite{HalvaniAssessingAVMethods:2019}, which are summarized below. 

%---------------------------------------------------------------------------------------------------------
The first category are \textbf{unary} \AV methods that determine their classification model solely on the basis of $\Arefset$. A unary \AV method assumes $\Dunk$ to be written by $\A$, if it is stylistically similar to the documents in $\Arefset$. 
%---------------------------------------------------------------------------------------------------------
The second category are \textbf{binary-intrinsic} \AV methods that determine their classification model on the basis of a given training corpus. This corpus consists of a number of verification cases with a ground truth of the form \classY (same-author) and \classN (different-author). A binary-intrinsic \AV method treats the unknown and known documents as a single unit $X$ (for example, a feature vector). If $X$ is more similar to the \classY-cases, the method accepts $\A$ as the author of $\Dunk$. If, on the other hand, $X$ is more similar to the \classN-cases, $\Dunk$ is assumed to be written by another author. In any case, the decision is made solely the basis of $X$ and the learned model (hence, intrinsic). 
%---------------------------------------------------------------------------------------------------------
The third category are \textbf{binary-extrinsic} \AV methods that determine their classification model on the basis of external (so-called \e{impostor} \cite{KoppelWinter2DocsBy1:2014}) documents which, for example, are gathered by using a search engine. In this context, the documents in $\Arefset$ represent samples of the class \classY, while the impostor documents act as samples of the counter class \classN. A binary-extrinsic \AV method assumes $\Dunk$ to be written by $\A$, if it is stylistically similar to the documents in $\Arefset$. Otherwise, $\A$ is rejected as the true author, if $\Dunk$ is more similar to the impostor documents. 
%---------------------------------------------------------------------------------------------------------
Over the last two decades, numerous \AV approaches have been proposed that can be assigned to one of these three categories. 
%---------------------------------------------------------------------------------------------------------

%---------------------------------------------------------------------------------------------------------
A recently published unary \AV approach, which we refer to as \nealAVIF, was developed by Neal et al. \cite{NealAVviaIsolationForests:2018} for the purpose of continuous verification. Their method is based on an isolation forest classifier, which, like many other \AV methods, considers \charNgrams as underlying features. \nealAVIF yielded a high recognition accuracy using very small training samples of 50 and 100-character blocks. However, in their study the authors explain that the method was only evaluated on positive samples (in other words, instances of the \classY-class). Therefore, it is not clear how well \nealAVIF performs under realistic conditions in which both classes (\classY \textbf{and} \classN) exist. 
%---------------------------------------------------------------------------------------------------------

%---------------------------------------------------------------------------------------------------------
A common binary-intrinsic \AV approach, which we denote by the name \stamatatosProf, was proposed by Potha and Stamatatos \cite{StamatatosProfileCNG:2014}. Their method considers two documents $\Dunk$ and $\DA$ as character $n$-gram profiles and measures their relative differences using a predefined dissimilarity function. If the resulting dissimilarity score exceeds a certain threshold (derived from the distribution of \classY/\classN-samples in a given training corpus), $\Dunk$ is assumed to be written by $\A$. Potha and Stamatatos \cite{StamatatosProfileCNG:2014} demonstrated that \stamatatosProf was able to outperform every single \AV method submitted to the first \AV-competition as a part of the PAN shared tasks \cite{PANOverviewAV:2013}. 
%---------------------------------------------------------------------------------------------------------

%---------------------------------------------------------------------------------------------------------
One of the most influential and successful binary-extrinsic \AV approach is the \e{Impostors Method} (\koppelIM) proposed by Koppel and Winter \cite{KoppelWinter2DocsBy1:2014}, which laid the foundations for many subsequent \AV approaches (for example,   \cite{SeidmanPAN13:2013,KhonjiIraqiAV:2014,KocherPANSpatium:2015,KocherSavoySpatiumL1:2017,StamatatosPothaImprovedIM:2017}). 
\koppelIM can be broken down into two steps. In the first step, appropriate impostor documents have to be collected according to a predefined strategy (\eg using a search engine or a static corpus with suitable documents). In the second step, a feature randomization technique is applied iteratively to measure the similarity between pairs of documents. If, given this measure, a suspect is picked out from among the impostor set with sufficient salience, then the suspect is assumed to be the author of $\Dunk$ \cite{KoppelWinter2DocsBy1:2014}. 
Two variants of \koppelIM, namely \khonjiGI proposed by Khonji and Iraqi \cite{KhonjiIraqiAV:2014} and \koppelGI proposed by Seidman \cite{SeidmanPAN13:2013} were the best-performing approaches in the first and second PAN-\AV competitions \cite{PANOverviewAV:2013,PANOverviewAV:2014}. 
%---------------------------------------------------------------------------------------------------------
An alternative binary-extrinsic \AV approach is the \veenmanNNCD method proposed by Veenman and Li \cite{VeenmanPAN13:2013}. In contrast to \koppelIM, their method delegates the entire feature engineering procedure to a \stateOfTheArt compression-algorithm. Here, $\Dunk$ is assumed to be written by $\A$ if the compressed version of $\Dunk$ is dissimilar to the compressed version of the impostor documents. Both \veenmanNNCD \cite{VeenmanPAN13:2013} and \koppelGI \cite{SeidmanPAN13:2013} were the best performing approaches in the first PAN-\AV competition \cite{PANOverviewAV:2013}. 
%---------------------------------------------------------------------------------------------------------

%% file: texfiles/ProposedApproach.tex
\section{Feature Categories} \label{Proposed_FeatureCategories} 
In this section, we propose a number of feature categories that are used by our \AV approach to capture the writing style of documents. A part of these derive from certain feature categories used in previous studies. The remaining feature categories, however, have been not considered so far in the context of \AV, at least to our best knowledge. All feature categories are summarized in Table~\ref{table:ProposedFeatureCategoriesWithSamples} along with a number of examples. In the following subsections, we first introduce all feature categories in detail. Afterwards, we explain which design decisions we made in regard to their hyperparameters. Finally, we describe the scope from where all proposed features are extracted and how we normalized them. 
\begin{table*} 
	\begin{center}\footnotesize
		\begin{tabular}{lllll} \toprule		
			%============================================================================================================
			\textbf{ID} & \textbf{Feature category}       & \textbf{Range}          & \textbf{Sample $\bm{n}$}  &  \textbf{Sample output} \\\midrule
			%============================================================================================================			
			$F_{1-3}$   & Punctuation $n$-grams           & $n \in \{1, 2, 3\}$     & $n=2$ & $\{ (\verb|'.| ) \}$     \\    
			$F_4$       & TA sentence and clause starters & \multicolumn{1}{c}{---} &       & $\{ (\verb|so| ) \}$   \\
			$F_5$       & TA sentence endings             & \multicolumn{1}{c}{---} &       & $\{ (\verb|goes| ) \}$     \\ 
			$F_{6-9}$   & TA token $n$-grams              & $n \in \{1, 2, 3, 4\}$  & $n=3$ & $\{ (\verb|so that's the| )\, , \; (\verb|it goes .| )	 \}$   \\
			$F_{10-11}$ & TA masked token $n$-grams       & $n \in \{3,4\}$         & $n=3$ & $\{ (\verb|that's the #| )\, , \; ( \verb|the # it|) \, , \; ( \verb|# it goes|)\}$   \\ \bottomrule			
			%============================================================================================================
		\end{tabular}
	\end{center}
	\caption{All 11 feature categories considered by \taveer (feature categories with the TA-prefix are proposed by us). The last column shows the output for the example sentence: \texttt{"So that's the way it goes."} Note that regarding the $\bm{n}$-grams, each setting of $\bm{n}$ results in an individual feature category.  \label{table:ProposedFeatureCategoriesWithSamples}} 
\end{table*}

\subsection{Topic-Agnostic Words and Phrases} \label{Proposed_FeatureCategories_TopicAgnosticPatterns} 
Function words can be seen as the most common choice in the field of authorship analysis, when it comes to select topic-agnostic features.  However, in the literature it often remains unclear what is exactly understood and represented under the term \quote{function words}. In many existing studies (for example, \cite{ChandrasekaranAAviaNN:2013,JuolaStolermanLingAuthAA:2013,ZhaoRelativeEntropyAA:2006}) no detailed explanation is provided regarding the question, which specific function word categories (or at least which specific words) were taken into account. Another peculiarity that can be seen in the literature, is the varying number of considered function words. For example, Chandrasekaran \cite{ChandrasekaranAAviaNN:2013}, Binongo \cite{BinongoAABookOfOz:2003}, Srinivasa \cite{SrinivasaAAImbalanced:2017} and Zhao and Zobel \cite{ZhaoEffectiveFunctionWordsAA:2005} make use of 24, 50, 150 and 365 function words, respectively. In view of these different numbers, the question arises why only individual subsets are considered rather than using the entire spectrum of function words. Instead of making use of non-structured and incomplete lists, Varela et al. \cite{VarelaAAVerbsAndPronouns:2010} and Pavelec et al. \cite{PavelecAAConjunctionsAndAdverbs:2008} follow a different approach, in which they consider subcategories of function words such as pronouns, conjunctions, subclasses of adverbs and other word forms. By this, a better insight can be gained regarding the question which specific type of function words were actually taken into account. 

Motivated by this idea, we opted for a similar but more systematic approach, in which we consider all existing categories of function words along with other carefully selected topic-agnostic (hereafter, abbreviated as \textbf{TA}) categories. First, we assemble a comprehensive list $\listStylePatterns$ consisting of words and phrases that belong to these categories (cf. Table~\ref{table:POSNoiseFeatures}). Based on $\listStylePatterns$, we then derive different TA feature categories (described below) that can be used to model the writing style of documents across different linguistic layers. For the construction of $\listStylePatterns$, we use a variety of words and phrases classified into 20 categories including function words, empty verbs, contractions, generic adverbs as well as transitional words and phrases. All considered words and phrases, which are known in the literature \cite{PavelecAAConjunctionsAndAdverbs:2008,BinongoAABookOfOz:2003,StolermanPhD:2015} to be content and topic independent, have been collected from different sources, in particular, linguistic books and stylometry papers. The transitional phrases cover a number of categories including \e{causation}, \e{contrast}, \e{similarity}, \e{clarification}, \e{conclusion}, \e{purpose} and \e{summary}. With regard to the verbs, we take the respective tenses\footnote{To generate the tenses, we used the \e{pattern} framework \cite{SmedtPatternPython:2012} available at \url{https://github.com/clips/pattern}.} into account (for example, \texttt{give} $\rightarrow \{ \texttt{gives, giving, gave, given} \}$) in order to enrich $\listStylePatterns$. All categories of words and phrases contained in $\listStylePatterns$ are summarized in Table~\ref{table:POSNoiseFeatures} along with a number of examples. 
\begin{table} 
	\centering\footnotesize	
	\begin{tabular}{ll} 
		\toprule 
		%============================================================================================================
		\textbf{Category} & \textbf{Examples} \\ \midrule
		%============================================================================================================
		Conjunctions         & $\{\texttt{and, as, because, but, either, for, hence, however, if, neither, nor, once,}\,\e{...}\,\}$  \\
		%------------------------------------------------------------------------------------------------------------ 		
		Determiners          & $\{\texttt{a, an, both, each, either, every, no, other, our, some,}\,\e{...}\,\}$  \\
		%------------------------------------------------------------------------------------------------------------ 		
		Prepositions         & $\{\texttt{above, across, after, among, below, beside, between, beyond, inside, outside,}\,\e{...}\,\}$  \\
		%------------------------------------------------------------------------------------------------------------		
		Pronouns             & $\{\texttt{all, another, any, anyone, anything, everything, few, he, her, hers, herself,}\,\e{...}\,\}$  \\
		%------------------------------------------------------------------------------------------------------------		
		Quantifiers          & $\{\texttt{any, certain, each, either, few, less, lots, many, more, most, much, neither,}\,\e{...}\,\}$  \\	\midrule	
		%============================================================================================================
		Auxiliary verbs      & $\{\texttt{can, could, might, must, ought, shall, will,}\,\e{...}\,\}$  \\
		%------------------------------------------------------------------------------------------------------------ 
		Delexicalised verbs  & $\{\texttt{get, go, take, make, do, have, give, set,}\,\e{...}\,\}$  \\
		%------------------------------------------------------------------------------------------------------------ 
		Empty verbs          & $\{\texttt{do, did, does, got, getting, have, had, had, gives, giving, gave, give, gets,}\,\e{...}\,\}$  \\
		%------------------------------------------------------------------------------------------------------------ 
		Helping verbs        & $\{\texttt{am, is, are, was, were, be, been, being, will, should, would, could,}\,\e{...}\,\}$  \\ \midrule
		%============================================================================================================ 
		Contractions         & $\{\texttt{i'm, i'd, i'll, i've, he's, it's, we'd, she's, it'll, we're, how's, you're,}\,\e{...}\,\}$  \\ \midrule
		%============================================================================================================	
		Adverbs of degree    & $\{\texttt{almost, enough, hardly, just, nearly, quite, simply, so, too,}\,\e{...}\,\}$  \\	
		Adverbs of frequency & $\{\texttt{again, always, never, normally, rarely, seldom, sometimes, usually,}\,\e{...}\,\}$  \\		
		Adverbs of place     & $\{\texttt{above, below, everywhere, here, in, inside, into, nowhere, out, outside, there,}\,\e{...}\,\}$  \\
		Adverbs of time      & $\{\texttt{already, during, immediately, just, late, recently, still, then, sometimes, yet,}\,\e{...}\,\}$  \\		
		Pronominal adverbs   & $\{\texttt{hereafter, hereby, thereafter, thereby, therefore, therein, whereas, wherever,}\,\e{...}\,\}$  \\
		Focusing adverbs     & $\{\texttt{especially, mainly, particularly, generally, only, simply, exactly, merely, solely,}\,\e{...}\,\}$  \\
		Conjunctive adverbs  & $\{\texttt{likewise, meanwhile, moreover, namely, nonetheless, otherwise, perhaps, rather,}\,\e{...}\,\}$  \\ \midrule
		%============================================================================================================
		Transition words     & $\{\texttt{besides, furthermore, generally, hence, thus, however, incidentally, subsequently,}\,\e{...}\,\}$  \\
		Transitional phrases & $\{\texttt{of course, as a result, in addition, because of, in contrast, on the other hand,}\,\e{...}\,\}$  \\
		Phrasal prepositions & $\{\texttt{as opposed to, in regard to, in relation to, inspite of, out of, with regard to,}\,\e{...}\,\}$  \\
		%------------------------------------------------------------------------------------------------------------				
		\bottomrule		
	\end{tabular}
	\caption{All categories of TA-based words and phrases. The list $\listStylePatterns$ is created by taking the union of all categories. \label{table:POSNoiseFeatures}}
\end{table}
Note that due to the ambiguities occurring in the English language, a number of function words appear in multiple categories. For example, \texttt{"but"} and \texttt{"for"} are both prepositions and conjunctions, whereas \texttt{"few"} represents a pronoun and a quantifier. However, regarding the features in $\listStylePatterns$, we do not differentiate between the different meanings of these homographs\footnote{Homographs are words with the same spelling but different meaning.}. Based on $\listStylePatterns$, we derive additional feature categories which are described in the following.

\subsubsection{Punctuation $n$-Grams $(F_{1-3})$} \label{Proposed_FeatureCategories_PunctuationNGrams}
Punctuation marks represent syntactic features that quantify the grammatical structures an author uses and, thus, are content and topic independent \cite{StolermanPhD:2015}. As punctuation $n$-grams we define a sequence of consecutive punctuation marks where letters, digits and other non-punctuation characters are skipped (cf. Table~\ref{table:ProposedFeatureCategoriesWithSamples}). Among others, punctuation $n$-grams capture specific symbols that occur at word-internal level such as hyphens or apostrophes used in contractions (\eg \texttt{"we\textbf{'}ve"} or \texttt{"they\textbf{'}re"}). Furthermore, they allow to recognize unusual punctuation habits reflecting the individual writing style of an author such as combinations of question and exclamation marks (\eg \texttt{"\textbf{?!?}"} or \texttt{"\textbf{!?!}"}), which occur in informal documents. In total, we consider three punctuation $n$-gram feature categories ($F_{1-3}$) that are not dependent on the list $\listStylePatterns$. However, the feature categories $F_{6-11}$ make use of $F_1$ (punctuation unigram).

\subsubsection{TA Sentence and Clause Starters $\boldmath{(F_{4})}$}
Words or phrases that appear at the beginning of sentences or clauses can reflect one aspect of an author's writing style. We therefore consider such sentences and clause starters as a distinct feature category. However, since our focus lies on TA-based features, we make sure that a word or phrase appearing at the beginning of a sentence or a clause is included in $\listStylePatterns$. Note that in case of clauses, we consider the preceding punctuation mark (comma or semicolon) together with the subsequent word or phrase as a whole feature (cf. Table~\ref{table:ProposedFeatureCategoriesWithSamples}).

\subsubsection{TA Sentence Endings $\boldmath{(F_{5})}$} \label{Proposed_FeatureCategories_SentenceEndings}
Words or phrases that appear at the end of sentences might also reflect a stylistic habit of authors. 
We therefore consider such features as a distinct feature category and make sure (analogous to $F_4$) that they are included in $\listStylePatterns$.

\subsubsection{TA Token $n$-Grams $\boldmath{(F_{6-9})}$} \label{Proposed_FeatureCategories_TATokenNGrams}
These feature categories can be seen as a form of standard token $n$-grams with the restriction that each token $t_i$ in a token $n$-gram $(t_1, t_2, \ldots, t_n)$ represents either a punctuation or a word appearing in $\listStylePatterns$ (cf. Table~\ref{table:ProposedFeatureCategoriesWithSamples}). Note that for $n = 1$, the respective feature category $F_6$ is essentially the list $\listStylePatterns$, which is obtained by merging all categories listed in Table~\ref{table:POSNoiseFeatures}.

\subsubsection{TA Masked Token $n$-Grams $\boldmath{(F_{10-11})}$}
These feature categories also represent a form of token $n$-grams with the restriction that $n - 1$ tokens in a token $n$-gram $(t_1, t_2, \ldots, t_n)$ are either punctuation marks or words appearing in $\listStylePatterns$. The remaining $n - 2$ tokens, on the other hand, represent topic-related words, which are then \textbf{masked} by the non-punctuation character \verb|#|. The intention behind these feature categories is to enable the detection of contexts surrounding or adjacent to topic-agnostic words (cf. Table~\ref{table:ProposedFeatureCategoriesWithSamples}).

\subsection{Feature Category Ranges} \label{FeatureCategories_HyperparameterRanges}
In previous \AV works (\eg \cite{StamatatosProfileCNG:2014,JankowskaAVviaCNG:2014,BrocardoStylometryAV:2013}) $n$-gram-based feature categories have been treated as a single concept, where the most suitable $n$ was chosen on the basis of a hyperparameter optimization procedure. In contrast to this, we treat $n$-gram-based feature categories \textbf{independently} so that, for example, punctuation $2$- and $3$-grams represent two individual feature categories. There is a simple justification for this decision: If we would restrict ourselves to only one specific $n$, optimized on a training corpus, we might miss important features occurring in the unseen data (test corpus) that can only be captured with an alternative setting of $n$. Allowing multiple settings of $n$ for the same feature category can, therefore, help to counteract a possible mismatch between training and test data. 

In the following, we explain the considerations behind the ranges of the $n$-gram-based feature categories listed in Table~\ref{table:ProposedFeatureCategoriesWithSamples}. For the punctuation $n$-grams, we set $n = 1$ as a lower limit which is useful in cases where sentences comprise only a single punctuation (\eg full-stop, question or exclamation mark). As an upper limit, we set $n = 3$, as it can be expected that longer punctuation sequences between the unknown and known documents will be scarce (more on this in the next subsection). 
%----------------------------------------------------------------------------------------------------------------
Regarding TA token $n$-grams, we set $n = 1$ and $n = 4$ as a lower and upper limit, respectively. For the former, we aim to capture at least single words in the documents. Here, we expect that a part of these features will be present in both documents, in most of the cases. With regard to longer sequences, we aim to capture specific phrases that can be relevant for individual authors. However, sequences with more than four tokens are less likely to appear, especially between short documents so that $n = 4$ can be seen as a good compromise. For the TA masked token $n$-grams, we set $n = 3$ as a lower limit, as one of our intentions is to capture (masked) topic words surrounded by topic-agnostic words, so that $n = 3$ is a minimum limit. As an upper limit, we set $n = 4$ for the same reason mentioned for TA token $n$-grams.

\subsection{Scope of Feature Extraction} \label{FeatureExtraction_Scope}
In existing \AV studies it is often not mentioned which \textbf{scope} is considered to extract $n$-gram-based features. Here, the scope might be the entire text, paragraphs, sentences, clauses, phrases or tokens. Depending on the considered scope, the dimension of the generated feature space may vary which, in turn, may affect the verification results. For example, extracting token $n$-grams from single sentences would result in a smaller number of features, in contrast to the extraction from the whole text. This is because token $n$-grams cross sentence boundaries, so that respective cross-sentence features are not taken into account. Despite the smaller number of available features, we have decided, with regard to our \AV approach, to extract all $n$-gram-based features exclusively from the \textbf{sentence-level} of the documents. The reason for this is that in practice short text fragments (\eg social media posts or email text bodies) are often concatenated to obtain a sufficient document length, so that one sentence might not always have a connection to a subsequent sentence. Hence, if we extract $n$-gram-based features from the entire text, we would erroneously create artificial cross-sentence features that may not occur in texts of a particular author. Note that for feature extraction, we only consider lower case in order to capture all possible case variants (for example, \texttt{"the"}, \texttt{"The"}, \texttt{"THE"}), which can occur especially in informal texts.

\section{Verification Method} \label{ProposedApproach}
In this section, we present our \AV approach \textbf{\textsf{TAVeer}}\footnote{\taveer stands for \quote{\textbf{T}opic-agnostic \textbf{A}uthorship \textbf{V}erifier based on \textbf{e}qual \textbf{e}rror \textbf{r}ate}.}, which is inspired by the methodology of biometric recognition systems. These aim to recognize individuals, based on a variety of physiological characteristics and behavioral features obtained from e.g. the hand, vein, fingerprint, face, eye, ear or voice. Here, the equal error rate (EER) represents a statistic used to show biometric performance in the context of a verification task. Essentially, EER corresponds to a point on a ROC curve where the false acceptance rate is equal to the false rejection rate. 

Given a questioned document $\Dunk$ and a document $\DA$ from a known author $\A$, the goal of our method is to determine whether $\Dunk$ was also written by $\A$. To achieve this goal, \taveer employs an ensemble of $m$ distance-based classifiers, where each one aims to accept or reject the questioned authorship of $\Dunk$. Each classifier is provided with a category of stylistic features extracted from an individual linguistic layer (in each document). In this context, EER serves as a thresholding mechanism, where erroneous verification predictions in either direction are treated equally. This is different from other \AV methods. The approach of Bevendorff et al. \cite{BevendorffUnmasking:2019}, for example, heavily prioritize precision over recall. 

\taveer can essentially be divided into the two phases \textbf{training} and \textbf{inference}. In the training stage, a model $\Model$ has to be \quote{learned} on the basis of a given training corpus $\Corpus = (\Problem_1, \Problem_2, \ldots, \Problem_n)$. Here, each $\Problem$ denotes a verification case, for which the ground truth (\classY/\classN) is known beforehand. In the inference stage, the generated model $\Model$ is applied to an unseen verification case $\Problem_{\e{?}}$ in order to accept or reject the questioned authorship. In what follows, we first describe the preliminaries for \taveer and afterwards its two stages.

\subsection{Preliminaries} 
Before describing our approach in detail, we first explain what exactly is considered as an input, how this input is represented and on which basic functionality it depends in order to measure the (dis)similarity between the documents.

\subsubsection{Document Input} 
\taveer follows the profile-based paradigm that, to our best knowledge, was first described by Potha and Stamatatos \cite{StamatatosProfileCNG:2014} in the context of \AV. Given a set of reference documents $\Arefset = \{ \D_1, \D_2, \, \ldots \}$ for a known author $\A$, the idea behind the profile-based approach is to concatenate all documents in $\Arefset$ into a single document $\DA$. Thus, a verification case $\Problem$ is transformed from $(\Dunk, \Arefset)$ to $(\Dunk, \DA)$, which represents the document input for \taveer.

\subsubsection{Document Representation}  \newcommand*{\vectorizer}{f}
As a document representation technique, we consider a \e{bag-of-features} model, in which all involved features are treated independently from each other. Let $\Fset = \{ F_1, F_2, \ldots F_m \}$ be the $m$ proposed feature categories (cf. Table~\ref{table:ProposedFeatureCategoriesWithSamples}) and $\Arefset$ be the global set of documents. We define a function $\vectorizer: \mathbb{D} \times \mathbb{D} \times \Fset \rightarrow \bigcup_{k \in \mathbb{N}} \mathbb{R}^k \times \mathbb{R}^k$, which transforms $\Dunk$ and $\DA$ according to a given feature category $F$ to two real valued vectors, where $k$ denotes the dimension of the feature space spanned by the features from $F$ contained in the documents $\Dunk$ and $\DA$. Consider for example $F_1$ as a feature category, which describes a set of punctuation marks $\{$\texttt{"-"}, \texttt{";"}, \texttt{"?", \e{...}}$\}$. Applying $\vectorizer$ to $\Dunk$ and $\DA$ yields all punctuation marks, that exist in at least one of the documents and adds them to a list $\mathcal{V} = (v_1, v_2, \ldots, v_k )$. Then, two vectors $X = (x_1, x_2, \ldots, x_k)$ and $Y = (y_1, y_2, \ldots, y_k)$ are created, where each $x_j$ and $y_j$ represents the absolute frequency of the corresponding punctuation mark $v_j \in \mathcal{V}$ in each document, respectively. As a final step, we normalize each vector by its \e{Manhattan norm}, denoted as $\|\cdot \|_{1}$, so that all contained features are scaled into the (real) interval $[0, 1]$ and sum up to one. This procedure holds for all $m$ feature categories.

\subsubsection{Distance Function}
To measure the (dis)similarity between two generated feature vectors $X$ and $Y$, we use a distance function dist$(X, Y)$. For this, we have chosen the well-known \e{Manhattan metric}, defined by:
\begin{equation}
\textrm{dist}(X,Y) = \| X - Y \|_1 = \sum_{r = 1}^{k} | X_r - Y_r | 
\label{eq:ManhattanDistanceFunction}
\end{equation} 
which has been used in a number of previous studies on stylometry (for example, \cite{BurrowsDelta:2002,AhmedAVFeatureCategories:2018}). The \e{Manhattan metric} benefits from its simplicity and also from the fact that it allows easy interpretation of which specific features have contributed to the verification result (as will be shown in Section~\ref{TAVeer_FeatureAnalysis}).

\subsection{Model Learning} \label{TAVeer_Model_Learning}
Given the training corpus $\Corpus$ and the set of all considered feature categories $\Fset = \{ F_1, F_2, \ldots F_m \}$, the objective of this step is to construct a model $\Model$, which represents the optimal combination of feature categories obtained on $\Corpus$. In the following, we describe the necessary sub-steps to create $\Model$.

\subsubsection{Computing Thresholds} 
In this sub-step, the individual thresholds $\Theta = (\Threshold_{F_1}, \Threshold_{F_2}, \ldots, \Threshold_{F_m})$ have to be computed for the $m$ feature categories. Using Equation~\ref{eq:ManhattanDistanceFunction}, we calculate for each verification case $\Problem_j = (\D_{\A, j}, \D_{\,\unknown, j}) \in \Corpus$ and each feature category $F_i$ the respective distance $d_{i, j} = \textrm{dist}(f(\D_{\A, j}, \D_{\,\unknown, j}, F_i))$. As a thresholding technique, we select the equal error rate (EER), which describes the point, where the false positives rate is equal to the false negatives rate. Since all corpora used in our experimental setting are balanced, a threshold, which will result in an EER, can be obtained by calculating the median of the distances over all cases in the corpus. Consequently, for all $m$ feature categories, we obtain the corresponding thresholds as follows:  
\begin{equation}
\Theta = ( \Threshold_{F_1}, \Threshold_{F_2}, \ldots, \Threshold_{F_m} )  \textrm{, with }  \Threshold_{F_i}  = \textrm{median}(d_{i, 1}, d_{i, 2}, \ldots, d_{i, n})
\label{eq:Thresholding}
\end{equation}
Note that in case where an exact EER is not feasible (for example, when multiple distance values are equal) the median provides the closest approximation of the EER.

\subsubsection{Similarity Function}
\begin{figure}
	\centering
	\begin{tikzpicture}[scale=2.8]	
	\draw[step=0.25, lightgray] (0,0) grid (2,1);	
	\foreach \x/\xtext in {0, 0.5, 1, 1.5, 2}
	\draw[shift={(\x,0)}] (0pt,1pt) -- (0pt,-1pt) node[below] {$\xtext$};
	\foreach \y/\ytext in {0, 0.5, 1}
	\draw[shift={(0,\y)}] (1pt,0pt) -- (-1pt,0pt) node[left] {$\ytext$};
	
	%\draw (1pt,1pt) -- (-1pt,-1pt) node[below left] {$0$};		
	
	\draw[->] (0,0) -- (2.2,0) node[right] {$d$};
	\draw[->] (0,0) -- (0,1.2) node[above] {$s$};	
	
	\draw[gray, purple] (0.3,0) -- (0.3,0.5);	
	\draw[gray, purple] (1.0,0) -- (1.0,0.5);	
	\draw[gray, purple] (1.8,0) -- (1.8,0.5);	
	
	\draw[gray, line width=0.75pt] (0,1) -- (0.3, 0.5) -- (2,0);
	\draw[gray, dashed, line width=0.75pt] (0,1) -- (1.0, 0.5) -- (2,0);
	\draw[gray, densely dotted, line width=0.75pt] (0,1) -- (1.8, 0.5) -- (2,0);
	
	\node at (0.3,0.5) [circle, fill=purple, minimum size=4pt,inner sep=0pt, outer sep=0pt,label=above:$\Threshold_{F_1}$] {};
	\node at (1.0,0.5) [circle, fill=purple, minimum size=4pt,inner sep=0pt, outer sep=0pt,label=above:$\Threshold_{F_2}$] {};
	\node at (1.8,0.5) [circle, fill=purple, minimum size=4pt,inner sep=0pt, outer sep=0pt,label=above:$\Threshold_{F_3}$] {};
	
	\end{tikzpicture} 	
	\caption{Behavior of the proposed similarity function with respect to the given distance $\bm{d}$ for $\bm{\dmax = 2}$ and three sample thresholds $\bm{\Threshold_{F_1} = 0.3, \Threshold_{F_2} = 1}$ and $\bm{\Threshold_{F_3} = 1.8}$. \label{fig:DistToSim}}		
\end{figure}
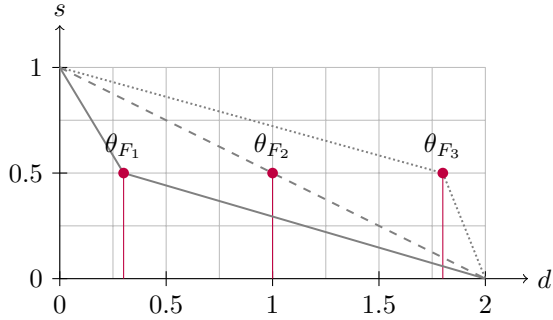

The introduced distance function (cf. Equation~\ref{eq:ManhattanDistanceFunction}) allows us to compute distances between pairs of feature vectors. However, the resulting distances are not calibrated with respect to the individual thresholds from the previous sub-step. Therefore, we designed a similarity function sim$(\cdot)$ that considers as an input a distance $d$, a threshold $\Threshold_{F}$ and the upper bound $\dmax$ of the provided distance function (in our case, the \e{Manhattan metric}). Recall that in the context of our approach, all feature vectors are normalized using the \e{Manhattan norm} $\|\cdot \|_{1}$. Consequently, all features in each vector sum up to 1. Based on this fact, the lower and upper bound of dist$(X, Y)$ can be calculated by  
\[
  \bm{0} \leq \| X - Y \|_1 \leq \| X \|_1 + \| Y \|_1  = \bm{2}
\]
such that $\dmax = 2$ holds. An important requirement regarding our similarity function is that the resulting score $s$ is calibrated in a way that 0.5 represents the decision boundary. One possible definition for a function sim$(\cdot)$ that transforms a distance $d$ into the range $[0, 1]$ and simultaneously calibrates the resulting similarity score $s$ with respect to this \quote{natural} decision boundary is: 
\begin{equation}
\textrm{sim}(d, \dmax, \Threshold_{F}) = \begin{cases}
1 - \frac{d}{2 \Threshold_{F}}, & \textrm{if}\;\, d \leq \Threshold_{F}, \\
\frac{1}{2} - \frac{d - \Threshold_{F}}  {2(d_{\e{max}} - \Threshold_{F})}, & \textrm{otherwise} \\
\end{cases} \label{eq:DistanceToSimilarity}
\end{equation} 
Figure~\ref{fig:DistToSim} illustrates the behavior of sim$(\cdot)$ with respect to the lower and upper bound of the \e{Manhattan metric}. 
Note that by considering $\dmax$ as a variable parameter, we can easily substitute the \e{Manhattan metric} with any other distance function, as long as its respective upper bound  $\dmax$ is known. Furthermore, it should be highlighted that any other definition for sim$(\cdot)$ that also fulfills the same requirement can be used instead.

\subsubsection{Classification Function}
The similarity function sim$(\cdot)$ from the previous sub-step can calculate a calibrated similarity value for a given distance $d$ and a threshold for a \textbf{single} feature category. However, the idea behind \taveer is to determine whether a questioned authorship between two documents holds based on \textbf{multiple} feature categories. Let $\Fset_{\Theta} = \{(F_i, \Threshold_{F_i}) | i \in \{1, 2, \ldots, m\} \}$ denote a set, which comprises pairs of feature categories and their associated thresholds and $\mathcal{P}(\Fset_{\Theta})$ the power set (without the empty set) holding all possible combinations of these pairs. We denote a single $\mathcal{E} \in \mathcal{P}(\Fset_{\Theta})$ by the term \textbf{ensemble}. Furthermore, we denote an ensemble comprising a single pair $\{ (F, \Threshold_F) \} \subseteq \Fset_{\Theta}$ as an \textbf{atomic ensemble}. To compute a similarity value with respect to $\mathcal{E}$, we define an aggregated similarity function sim$_\mathcal{E}(\cdot)$ as follows: 
\begin{equation}
\begin{aligned}
&\textrm{sim}_\mathcal{E}(\Dunk, \DA, \dmax, \mathcal{E}) = \textrm{median} \left( \{\textrm{sim}(\textrm{dist}(f(\Dunk, \DA, F)), \dmax, \Threshold_{F}) | (F, \Threshold_{F}) \in \mathcal{E} \} \right) 
\end{aligned} 
\label{eq:AggregatedSimilarityFunction}
\end{equation}

To obtain a binary prediction (\classY/\classN) for a single verification case $\Problem$ based on sim$_\mathcal{E}(\cdot)$, we further define a classification function:
\begin{equation}
\textrm{clf}(\Dunk, \DA, \dmax, \mathcal{E}) = \begin{cases}
\classY,\; \textrm{ if sim}_\mathcal{E}(\Dunk, \DA, \dmax, \mathcal{E}) > 0.5 \\
\classN,\; \textrm{ otherwise}
\end{cases}
\label{eq:ClassificationFunction}
\end{equation}

\subsubsection{Selecting Optimal Ensemble}
In this last sub-step, the goal is to determine the optimal ensemble, which will serve as the model $\Model$ for the inference stage, on the basis fo the training corpus $\Corpus$. To achieve this goal, we use Equation~\ref{eq:ClassificationFunction} to classify all verification cases $\Problem_1, \Problem_2, \ldots, \Problem_n$ in $\Corpus$ for each possible ensemble $\mathcal{E} \in \mathcal{P}(\Fset_{\Theta})$. As a result, we obtain $|\mathcal{P}(\Fset_{\Theta})|$ predictions for each $\Problem_i$. Based on the predictions and the ground truth provided for $\Corpus$, we can now calculate the accuracies for each ensemble to find the optimal one that will represent $\Model$. One way to obtain an optimal ensemble would be to select the one that leads to a maximum accuracy on $\Corpus$. In practice, however, this approach is not always reasonable as several ensembles can share the maximum accuracy. For this reason, we decided to consider additional criteria to obtain an optimal ensemble. Based on the power set $\mathcal{P}(\Fset_{\Theta})$, we sort all the resulting ensembles one by one according to the following three criteria (each in descending order): 
\begin{enumerate}
	\item Accuracy of an ensemble $\mathcal{E}$ (calculated for $\Corpus$)
	\item Number of feature categories an ensemble $\mathcal{E}$ contains 
	\item Median accuracy regarding all atomic ensembles in $\mathcal{E}$ (calculated for $\Corpus$)
\end{enumerate}
From here, it is unlikely that multiple ensembles share the same ranking regarding these criteria. Finally, we select the first ensemble from the sorted list, which will serve as the final model $\Model$.

\subsection{Inference} 
In contrast to the training phase, the inference phase is much more compact. Here, \taveer consumes the resulting model $\Model$ from the training phase and performs the following steps to classify an unseen verification case $\Problem_{\e{?}} = (\Dunk, \DA)$. Using Equation~\ref{eq:AggregatedSimilarityFunction}, \taveer first computes the similarity value $s_\e{?}$ between the unknown and known documents $\Dunk$ and $\DA$. Afterwards, a binary prediction regarding the questioned authorship of $\Dunk$ is obtained by comparing $s_\e{?}$ against the decision boundary 0.5 (cf. Equation~\ref{eq:ClassificationFunction}). In case that $s_\e{?} > 0.5$ holds, $\Problem_{\e{?}}$ is classified as \classY ($\Dunk$ and $\DA$ are assumed to be written by the same author), otherwise as \classN (both documents are probably written by different authors). 

%% file: texfiles/ProposedInterpretationScheme.tex
\section{Interpretation Method} \label{TAVeer_InterpretationScheme} 
Interpretability is a mandatory requirement in real forensic cases, as the classification result (a binary prediction and/or confidence score) of an \AV method alone is not sufficient to be used in legal proceedings. Rather, the actors involved (\eg a judge, a public prosecutor, an investigator and a suspect) must understand which particularities have influenced the decision of the \AV method under consideration. More precisely, it must be clear which specific features were involved in the analysis and how they contributed to the overall prediction of the \AV method. 
%----------------------------------------------------------------------------------------------------
Beyond the legal context, interpretability is also crucial in order to understand whether an \AV method is indeed focusing on the writing style of the questioned document and not accidentally on its topic or genre. Otherwise, it would indicate that the method does not fulfill its true purpose. 
%----------------------------------------------------------------------------------------------------
In the following, we present a simple technique that should contribute to a better understanding of this issue. Although the proposed approach was originally intended for \taveer, it can be adapted for other distance-based \AV methods as long as the specification described below is met. Note that the technique is intended for \e{post-hoc analysis} carried out by a human expert, \ie the investigator. This means that we assume that \taveer has already \textbf{seen} and \textbf{classified} the unknown document $\Dunk$. 
%----------------------------------------------------------------------------------------------------
%This allows the investigator to draw his/her own conclusions independently of the classification prediction of the \AV method. 
%----------------------------------------------------------------------------------------------------
%In Section~\ref{TAVeer_FeatureAnalysis} we provide an example for this. 

\begin{figure}[h!]
	\centering	
	\includegraphics[scale=0.56]{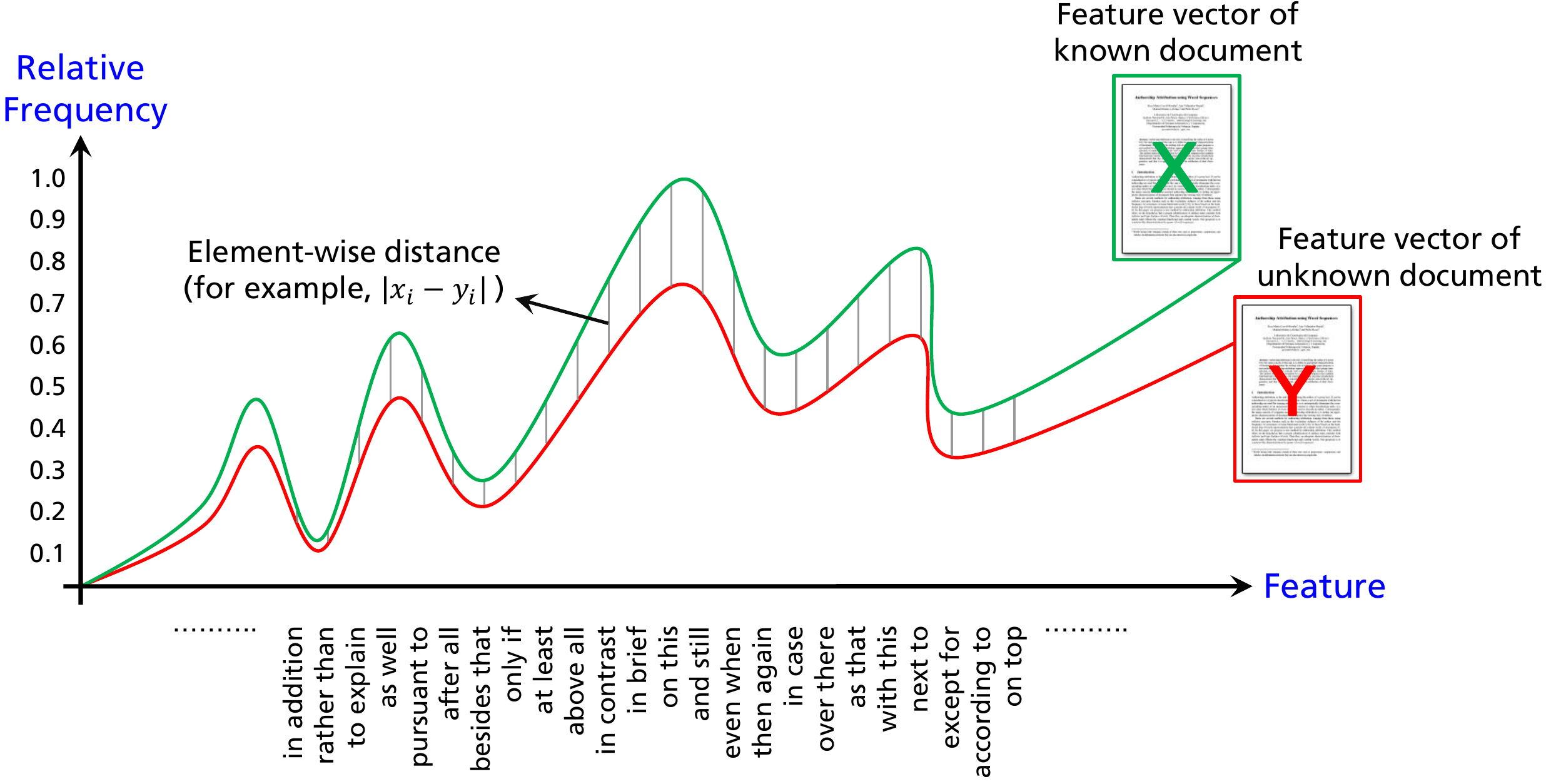}	
	\caption{Visualization of element-wise distances between both feature vector representations $X$ and $Y$ of the documents $\Dunk$ and $\DA$. Shorter distances between two features $x_i$ and $y_i$ contribute to a higher similarity between $X$ and $Y$ and vice versa. Note that the ``curves'' serve only for illustration purposes, since the horizontal axis represents discrete string-based features (here, features belonging to the feature category $F_7$ (TA token bigrams)) rather than continuous values. \label{figure:DistanceBasedInterpretability_NoThreshold}}	
\end{figure}

\begin{figure}[h!]
	\centering	
	\includegraphics[scale=0.55]{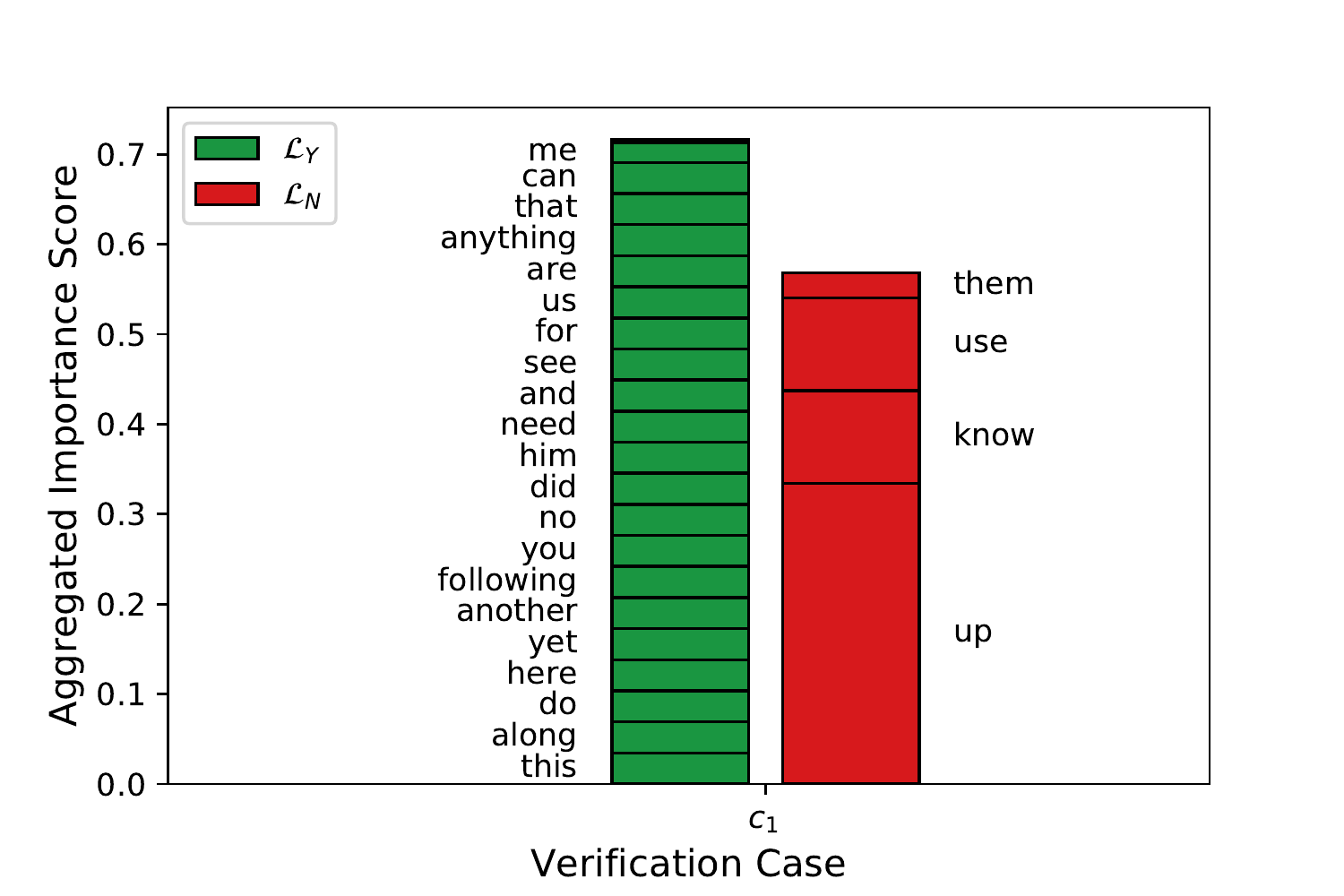}	
	\caption{Visualization of features belonging to the feature category $F_5$ (TA sentence endings) and their corresponding importance scores contained in \listY and \listN.  \label{figure:Interpretation_FeatureImportance_NoThreshold}}	
\end{figure}

The technique requires as an input a verification case $\Problem = (\Dunk, \DA)$, a distance function (in the context of this paper, the \e{Manhattan metric}), a specific feature category $F$ and a corresponding threshold $\Threshold_F$. The output is a triple $(\Phi, \textrm{\listY, \listN})$. Here, $\Phi$ represents a set of $m$ features, extracted from $\Dunk$ and $\DA$ using $F$, associated with their element-wise distances between $X$ and $Y$. The two lists $(\textrm{\listY, \listN})$, on the other hand, divide all $m$ features into disjoint partitions. The (string-based) features contained in \listY and \listN are associated with adjusted distance scores that aim to ``push'' \taveer towards a \classY- or an \classN-prediction, respectively. The entire procedure for generating $(\Phi, \textrm{\listY, \listN})$ is described in Algorithm~\ref{AVeerInterpretationScheme}. Once \listY and \listN have been created, we can use them to gain insight into what specific features have contributed to the decision of \taveer. 
%--------------------------------------------------------------------------------------------------------------- 
Inspecting $\Phi$, for example, allows a direct comparison (cf. Figure~\ref{figure:DistanceBasedInterpretability_NoThreshold}) regarding the question how (dis)similar the representations of $\Dunk$ and $\DA$ behave to each other with respect to the $m$ features and their corresponding element-wise distances. The decisive factor here is that the investigator is not ``distorted'' by the prediction of the \AV method, since the distances within $\Phi$ are not dependent on any threshold. Consequently, the investigator must draw her/his own conclusions regarding the authorship of $\Dunk$ on the basis of the features and their corresponding distances contained in $\Phi$. 
%--------------------------------------------------------------------------------------------------------------- 
With regard to the two lists \listY and \listN the situation is different. These consist of tuples of the form $(v, \featureImp)$, where $v$ represents a feature and $\featureImp$ its corresponding importance score so that the respective threshold is incorporated. Given these tuples, the investigator can see which features (and to what extent) contributed either to the \classY- or an \classN-prediction of the \AV method. 
%---------------------------------------------------------------------------------------------------------------
To illustrate the effects of the features contained in \listY and \listN, we visualize (cf. Figure~\ref{figure:Interpretation_FeatureImportance_NoThreshold}) them as vertically positioned rectangles, where green denotes 
\classY- and red \classN-scores. The height of each rectangle corresponds to the importance score $\featureImp$ of a feature $v$. Each $v$ (in the example provided in Figure~\ref{figure:Interpretation_FeatureImportance_NoThreshold}, a topic-agnostic word that occurs at the end of a sentence) is placed to the left or right of its respective rectangle. All \classY- and red \classN-rectangles are stacked atop each other. As a result, a given verification case $\Problem$ is represented by two stacked bars, so that $\Problem$ is classified as either \classY or \classN, depending on which stacked bar is higher. 
Figure~\ref{figure:Interpretation_FeatureImportance_NoThreshold} illustrates this idea. 
Here, we can see a verification case $\Problem_1$ taken from the test corpus $\CorpusReddit$ (cf. Section~\ref{Corpora_Reddit}), for which the features and importance scores contained in the generated lists \listY and \listN are plotted as a green and red stacked bar, respectively. $\Problem_1$ has been (correctly) classified as \classY, since the sum of all importance scores in \listY is bigger than the sum of the scores in \listN. 
%---------------------------------------------------------------------------------------------------------------
\begin{algorithm}
	\small
	%-----------------------------------------------------------------------------------------------------------------------
	\SetKwInput{KwData}{Input}
	\SetKwInput{KwResult}{Output}
	\SetKw{KwBreak}{break}
	\SetKw{KwContinue}{continue}
	\SetKw{KwThrow}{throw}
	\SetKwIF{If}{ElseIf}{Else}{if}{}{else if}{else}{end if}%
	\SetKwFor{For}{for}{}{end for}%
	\SetKwFor{ForEach}{foreach}{}{end foreach}%
	\SetKwFor{While}{while}{}{end while}%
	\SetNoFillComment
	\DontPrintSemicolon
	%-----------------------------------------------------------------------------------------------------------------------
	\KwData{Verification case $\Problem = (\Dunk, \DA)$, atomic ensemble $(F, \Threshold_{F})$ and a predefined distance function dist$(\cdot)$}
	\KwResult{Three lists $(\Phi, \textrm{\listY, \listN})$. The list $\Phi$ comprises features extracted via $F$ from the two documents $\Dunk$ and $\DA$ together with their associated distances. The two other lists \listY and \listN consist of the same features together with their importance scores.}
	%-----------------------------------------------------------------------------------------------------------------------
	\BlankLine
	
	\tcc{Given the feature category $F$, extract $m$ features that appear in $\Dunk$ or $\DA$.} 
	$\textsf{F} \leftarrow (f_1, f_2, \ldots, f_m)$
	\BlankLine
	
	\tcp{Construct normalized feature vectors for $\Dunk$ and $\DA$.}
	$X \leftarrow (x_1, x_2, \ldots, x_m)$ \\
	$Y \leftarrow (y_1, y_2, \ldots, y_m)$ 
	\BlankLine
	
	\tcc{Compute element-wise distances. For dist$(\cdot) =$ "Manhattan metric" an element-wise distance $d_i$ is computed by $|x_i - y_i|$} 
	$\textsf{D} \leftarrow (d_1, d_2, \ldots, d_m)$ 
	\BlankLine
	
	\tcp{Accociate each feature with its corresponding distance.}
	$\Phi \leftarrow \{ (v_1, d_1), (v_2, d_2), \ldots, (v_m, d_m) \}$ \\
	\BlankLine	
	
	\tcp{Define an equilibrated feature-wise threshold.}
	$\Threshold\,'_{F} \leftarrow \frac{1}{m}\Threshold_{F}$ \\
	\BlankLine
	
	\listY $\leftarrow (\,)$ \\ 
	\listN $\leftarrow (\,)$ \\
	\BlankLine 
	
	\ForEach{$(v_i, d_i) \in \Phi$} 
	{
		\tcc{Define importance score as the absolute difference between the feature-wise threshold and the distance.}
		$\featureImp_i \leftarrow |\Threshold\,'_{F} - d_i|$ \\ 
		\BlankLine
		\tcp{Assign feature and its importance score to \listY or \listN.}
		\If{$d_i < \Threshold\,'_{F}$}
		{
			\BlankLine
			Append $(v_i, \featureImp_i)$ to \listY \\
		}
		\Else
		{
			Append $(v_i, \featureImp_i)$ to \listN \\
		}
	}	
	\BlankLine	
	
	\KwRet{$(\Phi, \textrm{\listY, \listN})$} where \listY and \listN are sorted by the importance scores in descending order. 
	\caption{Interpretation Scheme for Distance-Based AV Methods}
	\label{AVeerInterpretationScheme}
\end{algorithm}

%% file: texfiles/Experiments.tex
\section{Experimental Evaluation} \label{Evaluation} 
This section gives a detailed description of our experimental evaluation. First, we introduce our self-compiled corpora and summarize their key statistics. Next, we describe which existing baseline methods we have selected to assess the performance of \taveer. Afterwards, we explain which performance measures we have chosen to evaluate all approaches. Finally, we present the results and describe our analytical findings.

\subsection{Corpora}
With regard to our experimental evaluation, we created \numberOfTestCorpora English corpora (cf. Table~\ref{tab:CorpusStatistics}) covering a variety of challenges such as documents with short lengths or cross-topic conditions between unknown and known documents. In total, the corpora comprise 7,178 verification cases, which were split into author-disjunct training and test sets based on a 40/60\% ratio. 
%In each corpus $\Corpus = \{ \Problem_1, \Problem_2, \ldots \}$, $\Problem_i$ denotes a verification problem that consists of a set of known documents $\Arefset$ and one unknown document $\Dunk$. 
For each known author in the \numberOfTestCorpora corpora, there is exactly one \classY- and one \classN-verification case. By this, we counteract the so-called \e{population homogeneity bias} described by Bevendorff et al. \cite{BevendorffBiasAV:2019}. Furthermore, we ensured that all corpora are \textbf{balanced} so that the number of \classY- and \classN-cases is equal. In the following, we present the four corpora in detail.  
\begin{table} [b]
	\centering\small 
	\begin{tabular}{lllrrrr}		
		\toprule
		%==================================================================================================
		\bfseries\boldmath Corpus $\Corpus$ & \textbf{Genre} & \textbf{Topic} & ~~~~\boldmath$|\Corpus|$ & ~~\boldmath$|\Arefset|$ & \bfseries\boldmath avg$|\DA|$ & \bfseries\boldmath ~~avg$|\Dunk|$  \\\midrule
		%==================================================================================================
		$\CorpusStackEx$ (train)   & Q$\,$\&$\,$A & Cross    & 150   & 1 & 11,247 & 9,956  \\
		$\CorpusStackEx$ (test)    & posts        & topics   & 228   & 1 & 11,803 & 10,700 \\\midrule
		%--------------------------------------------------------------------------------------------------------------				
		$\CorpusYelp$ (train)      & Restaurant   & Related  & 320   & 5 & 637    & 767 \\
		$\CorpusYelp$ (test)       & reviews      & topics   & 480   & 5 & 640    & 768 \\\midrule	
		%--------------------------------------------------------------------------------------------------------------
		$\CorpusReddit$ (train)	   & Social       & Mixed    &   800 & 3 & 5,735  & 6,785 \\
		$\CorpusReddit$ (test)     & news         & topics   & 1,200 & 3 & 5,854  & 6,794 \\\midrule
		%-----------------------------------------------------------------------------------------------------------
		$\CorpusAmazon$ (train)	   & Product      & Mixed    & 1,600 & 4 & 4,010  & 4,041 \\
		$\CorpusAmazon$ (test)     & reviews      & topics   & 2,400 & 4 & 4,009  & 4,041 \\
		%==================================================================================================
		\bottomrule	
	\end{tabular}
	\caption{Key statistics for our self-compiled corpora. Notation: $|\Corpus|$ denotes the number of verification cases in each corpus $\Corpus$, while $|\Arefset|$ denotes the number of the known documents. The average character length of $\Dunk$ and $\DA$ (concatenation of all documents in $\Arefset$) is denoted by avg$|\Dunk|$ and avg$|\DA|$, respectively. \label{tab:CorpusStatistics}}
\end{table}

\subsubsection{Stack Exchange Corpus ($\CorpusStackEx$)} \label{Corpora_StackExchange}
$\CorpusStackEx$ comprises 567 posts of 189 users, which have been crawled from the question-and-answer (Q$\,$\&$\,$A) network \e{Stack Exchange}\footnote{\url{https://stackexchange.com}} in 2019. The network comprises 173 Q$\,$\&$\,$A communities, where each community focuses on a specific topic. To construct $\CorpusStackEx$, we collected both questions and answers from users that were simultaneously active on the two thematically different communities \e{Cross Validated} ($\mathcal{S}_1$) and \e{Academia} ($\mathcal{S}_2$). $\CorpusStackEx$ is a strict \textbf{cross-topic} corpus, where each verification case comprises one known $\DA$ and one unknown document $\Dunk$. With regard to the \classY-cases, $\DA$ and $\Dunk$ stem from $\mathcal{S}_1$ and $\mathcal{S}_2$, respectively, while in each \classN-case, $\DA$ and $\Dunk$ stem from the same community $\mathcal{S}_1$. Consequently, inverse predictions (\ie \classN-cases are more likely to be classified as \classY and vice versa) can be expected for \AV methods that consider topic-related features, while topic-agnostic \AV methods are more likely to withstand the cross-topic characteristic of this corpus.

\subsubsection{Yelp Corpus ($\CorpusYelp$)}
$\CorpusYelp$ represents a collection of 2,400 comments written by 400 reviewers from the Yelp\footnote{\url{https://www.yelp.com/dataset/challenge}} business portal. This corpus is particularly demanding due to a number of peculiarities. First, it contains the shortest documents (compared to all other corpora), where each $\Dunk$ has a maximum length of ten sentences. Second, many reviews are written in different periods of time and often span $\approx4$ years, which results in a distorted writing style. Third, the reviews in $\CorpusYelp$ focus heavily on restaurants and food-related topics. Consequently, the number of relevant stylistic features is limited. Fourth, a part of the reviews contain a mixture of moderate and colloquial writing style covering a variety of slang expressions.

\subsubsection{Reddit Corpus ($\CorpusReddit$)}  \label{Corpora_Reddit}
$\CorpusReddit$ consists of 4,000 posts from 1,000 users, which were crawled between 2010--2016 from the community network \e{Reddit}. In contrast to the corpora $\CorpusStackEx$ and $\CorpusYelp$, $\CorpusReddit$ was designed as a \textbf{mixed-topic} corpus. For this, we collected for each author posts from various so-called \emph{subreddits}, where each subreddit focuses on a different topic. To construct a single verification case (\classY or \classN), we ensured that all contained documents were taken from disjoint \emph{subreddits} so that all documents are thematically different from each other. In total, $\CorpusReddit$ covers exactly 1,388 different topics including \e{politics}, \e{science}, and \e{movies}.

\subsubsection{Amazon Product Data Corpus ($\CorpusAmazon$)}  \label{Corpora_Amazon}
$\CorpusAmazon$ is derived from the \emph{Amazon product data} corpus, released by McAuley et al. \cite{AmazonReviewCorpus:2015}. The original dataset contains 142.8 million product reviews from the online-marketplace   \emph{Amazon}, gathered between 1996--2014. From this dataset, we extracted a subset of 10,000 reviews from 2,000 users, where each review is related to one of 17 product categories (\eg \e{electronics}, \e{movies and TV} or \e{office products}). Similarly to the $\CorpusReddit$ corpus, we made sure that all documents in each verification case differ from each other with respect to the product categories. Hence, $\CorpusAmazon$ also represents a mixed-topic corpus.

%============================================================================================================== 
\subsection{Existing Baselines} 
To measure the effectiveness of \taveer, we have selected \numberOfExistingBaselines competitive \AV methods as baselines, which have shown their strengths in a number of previous studies. Six of these methods (\genUnmask \cite{BevendorffUnmasking:2019}, \baff \cite{BevendorffBiasAV:2019}, \occav \cite{HalvaniOCCAV:2018}, \hossamAV \cite{AhmedAVFeatureCategories:2018}, \coav \cite{HalvaniARES:2017} and \spatium \cite{KocherSavoySpatiumL1:2017}) were proposed in the last three years, while the remaining four methods (\stamatatosProf \cite{StamatatosProfileCNG:2014}, \koppelIM \cite{KoppelWinter2DocsBy1:2014}, \veenmanNNCD \cite{VeenmanPAN13:2013} and \koppelUnmask \cite{KoppelAVOneClassClassification:2004}) were published earlier. In the following, we describe some design decisions we have made with regard to these methods in order to apply them in our evaluation.

\subsubsection{Source of Impostor Documents}
\koppelIM, \veenmanNNCD and \spatium represent extrinsic \AV methods and, thus, make use of external documents to transform a verification case from a unary to a binary classification problem. In the original papers, both \koppelIM and \veenmanNNCD make use of search engines to generate the impostor documents. However, due to quota limitations, we opted for an alternative strategy with regard to our reimplementations, where the impostor documents were directly taken from the test corpora. This strategy has been also considered by Kocher and Savoy \cite{KocherPANSpatium:2015,KocherSavoySpatiumL1:2017} for \spatium. Although using static corpora is not as flexible as using search engines, it has the advantage that due to the available metadata (for instance, user names of the authors) the true author of the unknown document is likely not among the impostors\footnote{However, we cannot guarantee if different user names in fact refer to different persons. In other words, it might be possible that multiple accounts refer to the same person.}.

\subsubsection{Uniform (Binary) Predictions} 
In their original form, \genUnmask, \spatium and \stamatatosProf allow the three possible prediction outputs \classY (same-author), \classN (different-author) and \unanswered (unanswered), whereas for the remaining approaches only binary predictions (\classY/\classN) are considered. Therefore, to enable a fair comparison,  we decided to unify the predictions of all involved \AV methods to the binary case. In this context, verification cases for which the \AV methods determined similarity values greater than 0.5 were classified as \classY, otherwise as \classN. Here, all similarity values were normalized into the range $[0, 1]$, so that 0.5 marks the decision threshold.

\subsubsection{Settings for the Compression-Based Methods}
For the compression-based \AV methods \veenmanNNCD, \coav and \occav we used the same compression algorithm PPMd as mentioned in the original papers  \cite{HalvaniOCCAV:2018,HalvaniARES:2017,VeenmanPAN13:2013}. However, in the respective papers it has not been mentioned how the hyperparameter \e{model-order} of PPMd has been set. We therefore decided to set this hyperparameter to 7 for all three methods, based on our observation\footnote{With regard to the hyperparameter \e{model-order}, we experimented with values from 2 to 10.} that this value led to the best accuracy across all our training corpora. Moreover, we used as dissimilarity functions CDM$(\cdot)$ for \veenmanNNCD as well as CBC$(\cdot)$ for \coav and \occav, as defined in the original papers. Apart from these, there are no other hyperparameters for these approaches.

\subsubsection{Counteracting Non-Deterministic Behavior}
The four \AV methods \spatium, \koppelIM, \koppelUnmask and \genUnmask involve different sources of randomness (\eg feature subsampling, chunk generation or impostor selection) and, due to this, cause non-deterministic behavior regarding their predictions. In other words, applying these methods multiple times to the same verification case can result in different prediction outputs \ie (\classY, \classN, \classY, \ldots, \classN) which, in turn, can lead to a biased evaluation. To counteract this problem, we performed 11 runs for each non-deterministic method and selected the run (together with the calculated accuracy, \auc and the four confusion matrix outcomes) for which the accuracy score represented the median. The reason why we avoided to average the multiple runs (as, for example, was the case in \cite{StamatatosPothaImprovedIM:2017}) was to obtain accurate numbers in regard to our analysis.

\subsubsection{Model and Hyperparameters}
In general, \textbf{model parameters} refer to parameters that are estimated directly from the data, while \textbf{hyperparameters} cannot be obtained directly from the data and must therefore be set manually. In regard to the \AV methods considered in our experiments, model parameters represent the weights that form the SVM-hyperplanes (used by \koppelUnmask and \genUnmask) or scalar thresholds required to accept or reject the questioned authorships (used by \coav, \stamatatosProf, \koppelIM and \hossamAV). To obtain the model parameters of \stamatatosProf, \koppelIM, \coav and \hossamAV, we trained the methods on the respective training corpora. 

The hyperparameters involved in our selected \AV approaches represent, among others, the number of $k$ cross-validation folds (used by \koppelUnmask and \genUnmask) or the $n$-order of the character $n$-grams (used by \stamatatosProf and \hossamAV) and have been tuned in the following way: For \spatium, we used the original implementations\footnote{Both implementations are available at \url{https://github.com/pan-webis-de}.} together with their unmodified hyperparameter settings, mentioned in the respective papers \cite{KocherPANSpatium:2015,KocherSavoySpatiumL1:2017}. Regarding \stamatatosProf, \koppelIM, \koppelUnmask, \genUnmask and \hossamAV, we used our own implementations, where for the first two we employed the same hyperparameter ranges described in the original papers. For \genUnmask, we have considered the same fixed\footnote{These are: \e{Number and size of chunks (in words):} 30 chunks counting 700 words each, \e{Initial feature set size:} 250 most frequent words in each document pair, \e{Number of cross-validation folds:} 10, \e{SVM-Kernel:} linear, \e{Number of eliminated features:} 5 (positive and negative, respectively).} hyperparameters mentioned in the original paper \cite{BevendorffUnmasking:2019}, whereas for \koppelUnmask an adjustment was needed to fit our experimental setting. In the original definition of this method, Koppel and Schler \cite{KoppelAVOneClassClassification:2004,KoppelUnmasking:2007} used entire books to train and evaluate \koppelUnmask, which differ in lengths from the documents used in our corpora. Therefore, instead of using the original fixed hyperparameter settings (which would make \koppelUnmask inapplicable in our evaluation setting), we decided to consider individual hyperparameter ranges with values that are more appropriate for shorter documents as available in our corpora. The customized ranges are listed in Table~\ref{tab:UnmaskingHyperparams}. 
\begin{table} 
	\centering\small
	\begin{tabular}{lrr} 
		\toprule
		%=================================================================================================
		\textbf{Hyperparameter}       & \textbf{Our grid search range} & \textbf{Original setting}  \\ \midrule
		%=================================================================================================	
		% Official Unmasking Parameters.
		%--------------------------------------------
		$U_1$ = Initial feature set sizes     & $\{ 5, 15, 25, 35, 50, 75, 100, 150 \}$ & 250 \\
		$U_2$ = Number of eliminated features & $\{ 2, 3, 5 \}$ & 3 \\ 
		$U_3$ = Number of iterations          & $\{ 3, 5, 7 \}$ & 10 \\
		%--------------------------------------------
		% Unofficial Unmasking Parameters. 
		%--------------------------------------------
		$U_4$ = Chunk sizes (in words)        & $\{ 5, 15, 25, 35, 50, 75 \}$ & 500 \\
		$U_5$ = Number of folds               & $\{ 3, 5, 7, 10 \}$ & 10 \\
		%=================================================================================================	
		\bottomrule			
	\end{tabular}
	\caption{Adjusted hyperparameter ranges for the \AV method \koppelUnmask. Note that the most important modifications affect the hyperparameters $U_1$ and $U_4$.  \label{tab:UnmaskingHyperparams}}
\end{table}

Based on the original and adjusted ranges, \stamatatosProf, \koppelIM and \koppelUnmask were optimized using the grid search algorithm, which was guided by accuracy as a performance metric. All tuned hyperparameters are listed in Table~\ref{tab:BaselinesHyperparameters}. For a more detailed explanation of each hyperparameter, we refer the interested reader to the original paper of the respective \AV method. 
\begin{table} 
	\centering\small
	\begin{tabular}{p{0.21cm}lrrrr} 
		\toprule
		%=================================================================================================
		& \textbf{Hyperpar.} & $\bm{\CorpusStackEx}$ & $\bm{\CorpusYelp}$ & $\bm{\CorpusReddit}$ & $\bm{\CorpusAmazon}$ \\ \midrule		
		%-------------------------------------------------------------------------------------------------
		& $L_u$ & 		     9,000 & 		9,000 & 		   7,000 & 			 6,000 \\
		& $L_k$ & 			 5,000 &        2,000 & 		   9,000 &  		 1,000 \\ 
		& $n$   &				 4 & 			5 & 			   5 &				 5 \\
		\multirow{-4}{*}{\footnotesize\rotatebox{90}{\stamatatosProf}}
		& $d$   & 			 $d_1$ &        $d_0$ &            $d_0$ &           $d_1$ \\\midrule
		%-------------------------------------------------------------------------------------------------
		& $U_1$ &              150 &          100 &               50 &             150 \\
		& $U_2$ &                5 &            2 &                2 &               2 \\ 
		& $U_3$ &                5 &            7 &                7 &               3 \\ 
		& $U_4$ &                5 &            5 &                5 &               5 \\ 
		\multirow{-5}{*}{\footnotesize\rotatebox{90}{\koppelUnmask}}
		& $U_5$ &               10 &            7 &                7 &               7 \\\midrule	  
		%-------------------------------------------------------------------------------------------------			  
		&   $M$ &              100 &          100 &              100 &             100 \\
		&   $N$ &               10 &           10 &              100 &              10 \\
		\multirow{-3}{*}{\footnotesize\rotatebox{90}{\koppelIM}}
		&   $k$ &               50 &          100 &              100 &              50 \\\midrule
		%-------------------------------------------------------------------------------------------------			
		& $F$ & 		     Token & 		Lemma & 		   Token & 			 Token \\
		& Top $x$\% &           15 &	        8 &                1 & 		         3 \\ 
		\multirow{-3}{*}{\footnotesize\rotatebox{90}{\textsf{Dyn.AV}}}
		& $n$ &                  4 &		    1 & 	           1 & 			     1 \\\bottomrule		
		%-------------------------------------------------------------------------------------------------					
	\end{tabular}
	\caption{Hyperparameters of \stamatatosProf, \koppelUnmask, \koppelIM and \hossamAV tuned on the respective training corpora. Notation: 
		\fbox{\stamatatosProf} $L_u$ = {Profile size of the unknown document}; $L_k$ = {Profile size of the known document}; $n$ = {The $n$-order of character n-grams}; $d$ = {Dissimilarity function} (cf. \cite[Section~3.1]{StamatatosProfileCNG:2014}); \fbox{\koppelIM} $M$ = {Number of most similar documents that serve as potential impostors}; $N$ = {Number of actual impostors from among the potential impostors}; $k$ = {Number of iterations}; \fbox{\hossamAV} Top $x$\% frequently occurring $F$ = feature category $n$-grams; \fbox{\koppelUnmask} $U_1\,$--$\,U_5$ as described in Table~\ref{tab:UnmaskingHyperparams}. \label{tab:BaselinesHyperparameters}}
\end{table}

\subsubsection{Consideration of Features}
With regard to the \hossamAV approach, we have discarded the feature category \e{diacritics}, based on the observation that they were very rarely present within the documents in our corpora. With regard to the \posTags used by the same method, we used the well-known \e{spaCy}\footnote{Here, we used \e{spaCy}'s integrated POS tagger (model: \texttt{"en\_core\_web\_lg"}) available at \url{https://spacy.io}.} framework \cite{HonnibalJohnson:2015:EMNLP}.

\subsection{Performance Measures} 
To assess the performance of \taveer, we selected accuracy as a primary measure for a number of reasons. First, accuracy has been used in numerous research works including  \cite{AhmedAVFeatureCategories:2018,BarbonAV4CompromisedAccountsSocialNetworks:2017,BenzebouchiAVwithWordEmbeddings:2018,CastroAVAverageSimilarity:2015,ChenAuthorshipSimilarityDetection:2011,CastanedaAVviaLDA:2017,KoppelAVOneClassClassification:2004,KestemontUnmaskingAV:2012,StavngaardAVGhostwritingDetection:2019} and, thus, can be seen as the most common choice in the field of \AV. Second, the measure is intuitively understandable (the higher the resulting accuracy value, the lower the sum of incorrect predictions). Third, accuracy behaves symmetric in contrast to alternative performance measures (for example, \fOne) so that the classes \classY and \classN are treated equally. Fourth, the measure is suitable for our purpose, since in our experiments all corpora are balanced. 

For a better comparability, we also report the four confusion matrix outcomes: true positives (TP), false negatives (FN), false positives (FP) and true negatives (TN) that aim to provide a detailed insight into the predictions of the individual \AV methods. Moreover, they allow the interested reader to compute other performance measures that might be useful for other comparisons. Moreover, we consider \auc as an alternative measure, which has been often used in previous \AV studies including  \cite{AdamovicLanguageIndependentAV:2019,CastanedaAVviaLDA:2017,StamatatosPothaImprovedIM:2017,PothaStamatatosDynamicEnsembleAV:2019,PANOverviewAV:2015}.

\subsection{Results} 
After we have trained \taveer and the \numberOfExistingBaselines selected baselines on the \numberOfTestCorpora training corpora, we evaluated all \AV methods on the respective test corpora. 
The evaluation results are shown in Table~\ref{tab:ExperimentComparisonResults}. As can be seen in this table, \taveer outperforms all \numberOfExistingBaselines baselines on the demanding corpus $\CorpusStackEx$, which demonstrates its robustness under cross-topic conditions. In addition, \taveer surpasses all baseline methods on $\CorpusAmazon$ and is ranked second on $\CorpusReddit$, indicating that the method is also suitable for mixed-topic corpora. On the corpus $\CorpusYelp$, \taveer is on par with the strongest baseline $\koppelIM$, which shows that the method is also applicable in scenarios where the documents consist of a few sentences. It should be emphasized, that half of the baseline methods ($\koppelIM, \stamatatosProf, \occav$ and $\veenmanNNCD$) make use of \charNgrams which, according to the literature, represent the strongest features in the field of \AV. 
In this regard, the results demonstrate that punctuation- and TA-based features alone, are similarly effective and in some cases even better.  

Nevertheless, the accuracies listed in Table~\ref{tab:ExperimentComparisonResults} reflect only a summary view of \taveer's performance. To gain an insight into what led to these results, we therefore conduct a more detailed analysis regarding our approach. For this purpose, we first take a closer look at the models \taveer learned on the basis of the training corpora, in order to understand their behavior on the test corpora. Using the proposed interpretation technique (cf. Section~\ref{TAVeer_InterpretationScheme}) we then perform a more fine-grained analysis in which we investigate which specific features contributed to \taveer's predictions. 
\setlength{\aboverulesep}{0pt}
\setlength{\belowrulesep}{0pt}
\setlength{\extrarowheight}{.25ex}
\definecolor{lightgray}{rgb}{0.85, 0.85, 0.85}
\newcolumntype{g}{>{\columncolor{lightgray}}r}
\begin{table}
	\centering\small
	\begin{adjustbox}{max width=\textwidth}		
		\begin{tabular}{p{0.21cm}lrrrrrr}  % "r" mit "g" ersetzen um die background-farbe zu erhalten !
			\toprule
			%===========================================================================================================================================
			&  \textbf{Method} & \textbf{Acc.} & \textbf{AUC} & \textbf{TP} & \textbf{FN} & \textbf{FP} & \textbf{TN} \\\midrule			
			%=================================================================================================================================================
			% STACK EXCHANGE
			%=================================================================================================================================================
			& \cellcolor{LightGray}\taveer & \cellcolor{LightGray}\textbf{0.697} & \cellcolor{LightGray}0.778 & \cellcolor{LightGray}80 & \cellcolor{LightGray}34 & \cellcolor{LightGray}35 & \cellcolor{LightGray}79 \\
			& \coav & 0.404 & 0.388 & 42 & 72 & 64 & 50 \\
			& \koppelIM & 0.482 & 0.515 & 38 & 76 & 42 & 72  \\
			& \baff & 0.531 & 0.545 & 44 & 70 & 37 & 77 \\
			& \hossamAV & 0.496 & 0.518 & 87 & 27 & 88 & 26 \\
			& \veenmanNNCD & 0.513 & 0.552 & 4 & 110 & 1 & 113  \\
			& \occav & 0.496 & 0.408 & 0 & 114 & 1 & 113  \\
			& \stamatatosProf & 0.539 & 0.609 & 67 & 47 & 58 & 56  \\
			& \spatium & \underline{0.636} & 0.723 & 49 & 65 & 18 & 96  \\
			& \genUnmask & 0.522 & 0.524 & 54 & 60 & 49 & 65 \\
			\multirow{-10}{*}{\large\rotatebox{90}{$\bm{\CorpusStackEx}$}}
			& \koppelUnmask & 0.539 & 0.542 & 60 & 54 & 51 & 63  \\ \midrule	
			%=================================================================================================================================================
			% YELP
			%=================================================================================================================================================
			& \cellcolor{LightGray}\taveer & \cellcolor{LightGray}0.690 & \cellcolor{LightGray}0.746 & \cellcolor{LightGray}166 & \cellcolor{LightGray}74 & \cellcolor{LightGray}75 & \cellcolor{LightGray}165 \\
			& \coav & \textbf{0.710} & 0.769 & 166 & 74 & 65 & 175 \\
			& \koppelIM & \underline{0.708} & 0.788 & 150 & 90 & 50 & 190 \\
			& \baff & 0.592 & 0.704 & 206 & 34 & 162 & 78 \\
			& \hossamAV & 0.608 & 0.663 & 178 & 62 & 126 & 114 \\
			& \veenmanNNCD & 0.629 & 0.986 & 62 & 178 & 0 & 240 \\
			& \occav & 0.629 & 0.703 & 190 & 50 & 128 & 112 \\
			& \stamatatosProf & 0.665 & 0.723 & 155 & 85 & 76 & 164 \\
			& \spatium & 0.590 & 0.651 & 93 & 147 & 50 & 190 \\
			& \genUnmask & 0.500 & 0.500 & 0 & 240 & 0 & 240 \\
			\multirow{-10}{*}{\large\rotatebox{90}{$\bm{\CorpusYelp}$}}
			& \koppelUnmask & 0.596 & 0.639 & 153 & 87 & 107 & 133 \\ \midrule			
			%=================================================================================================================================================
			% REDDIT 
			%=================================================================================================================================================
			& \cellcolor{LightGray}\taveer & \cellcolor{LightGray}0.806 & \cellcolor{LightGray}0.861 & \cellcolor{LightGray}455 & \cellcolor{LightGray}145 & \cellcolor{LightGray}88 & \cellcolor{LightGray}512 \\
			& \coav & \textbf{0.836} & 0.909 & 503 & 97 & 100 & 500 \\
			& \koppelIM & \underline{0.833} & 0.888 & 431 & 169 & 31 & 569  \\
			& \baff & 0.759 & 0.824 & 422 & 178 & 111 & 489 \\
			& \hossamAV & 0.770 & 0.820 & 511 & 89 & 187 & 413 \\
			& \veenmanNNCD & 0.773 & 0.999 & 328 & 272 & 0 & 600  \\
			& \occav & 0.778 & 0.851 & 409 & 191 & 75 & 525  \\
			& \stamatatosProf & 0.764 & 0.821 & 453 & 147 & 136 & 464  \\
			& \spatium & 0.797 & 0.863 & 446 & 154 & 90 & 510 \\
			& \genUnmask & 0.585 & 0.621 & 328 & 272 & 226 & 374 \\
			\multirow{-10}{*}{\large\rotatebox{90}{$\bm{\CorpusReddit}$}}
			& \koppelUnmask & 0.719 & 0.785 & 467 & 133 & 204 & 396 \\	\midrule			%=================================================================================================================================================
			% AMAZON
			%=================================================================================================================================================
			& \cellcolor{LightGray}\taveer & \cellcolor{LightGray}\textbf{0.842} & \cellcolor{LightGray}0.912 & \cellcolor{LightGray}982 & \cellcolor{LightGray}218 & \cellcolor{LightGray}161 & \cellcolor{LightGray}1039 \\			
			& \coav & 0.768 & 0.847 & 925 & 275 & 281 & 919 \\
			& \koppelIM & \underline{0.815} & 0.901 & 941 & 259 & 186 & 1014  \\
			& \baff & 0.698 & 0.762 & 647 & 553 & 171 & 1029 \\
			& \hossamAV & 0.785 & 0.876 & 1030 & 170 & 346 & 854 \\
			& \veenmanNNCD & 0.600 & 0.996 & 239 & 961 & 0 & 1200 \\
			& \occav & 0.738 & 0.810 & 950 & 250 & 378 & 822  \\
			& \stamatatosProf & 0.723 & 0.797 & 861 & 339 & 326 & 874 \\
			& \spatium & 0.788 & 0.873 & 841 & 359 & 150 & 1050 \\
			& \genUnmask & 0.563 & 0.598 & 662 & 538 & 511 & 689 \\
			\multirow{-10}{*}{\large\rotatebox{90}{$\bm{\CorpusAmazon}$}}
			& \koppelUnmask & 0.725 & 0.801 & 903 & 297 & 362 & 838 \\	
			%=================================================================================================================================================
			\bottomrule
		\end{tabular}
		\end{adjustbox}
	\caption{Evaluation results of \taveer and the \numberOfExistingBaselines selected baseline methods. Bold and underlined values represent the best and second best results. \label{tab:ExperimentComparisonResults}}
\end{table}

\subsubsection{Model Analysis} \label{TAVeer_ModelAnalysis}
\begin{figure*}  % trim=[left bottom right top]    	% Für Debug-Zwecke --> \fbox{ }
	\centering
	\includegraphics[width=1.0\linewidth,trim=5.3cm 0.28cm 5.5cm 0.2cm,clip]{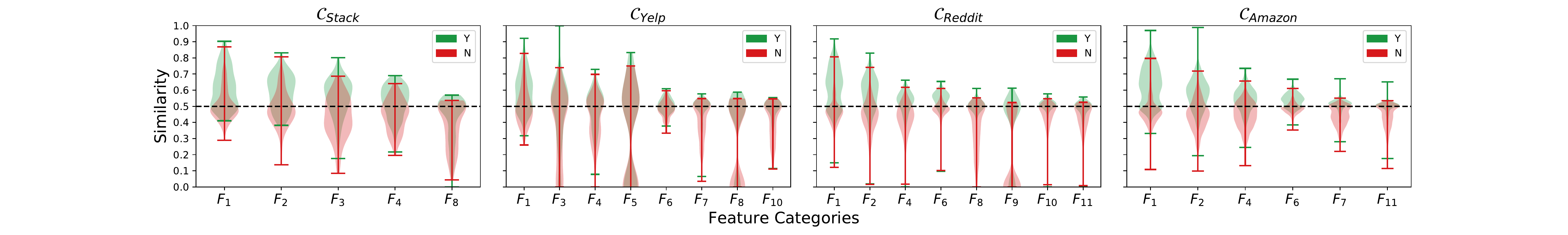}
	\caption{Similarity scores computed for all verification cases in the test corpora. The feature categories correspond to those in the models, learned on the respective training corpora (cf. Table~\ref{tab:OptimalFeatureCategories}). \label{ModelAnalysis_IndividualStrengthsOfFeatureCategories}}
\end{figure*}

The models learned by \taveer, using the procedure described in Section~\ref{TAVeer_Model_Learning}, contain some useful details that can help to understand the question, which feature categories are more relevant for each corpus. In the following, we take a closer look on the \numberOfTestCorpora generated models (cf. Table~\ref{tab:OptimalFeatureCategories}) to answer this question. As a starting point for our investigation, we apply \taveer using a model $\Model_{q}$ to a test corpus $\Corpus_{q}$ for each $q \in \{$Stack, Yelp, Reddit, Amazon$\}$. As a result, we obtain for each $\Corpus_{q}$ the predictions for all contained verification cases. Using Equation~\ref{eq:AggregatedSimilarityFunction}, we then compute similarity scores for all verification cases with respect to each atomic ensemble $\{(F, \Threshold_F)\} \subseteq \Model_{q}$. The similarity scores for the \classY- and \classN-cases in each $\Corpus_{q}$, regarding the feature categories contained in each model $\Model_{q}$, are visualized as violin plots in Figure~\ref{ModelAnalysis_IndividualStrengthsOfFeatureCategories}. These plots can be interpreted as follows. The distribution of the similarity scores for each feature category $F$ are colored green and red, respectively, while the dashed line represents the decision boundary. The better this line can separate both distributions and the less they overlap, the more suitable is $F$ for the test corpus $\Corpus_{q}$. 
%-----------------------------------------------------------------------------------------------------
Although no single feature category clearly separates the distributions across all corpora, we can still observe a number of tendencies. Regarding $\CorpusStackEx, \CorpusReddit$ and $\CorpusAmazon$, for example, it can be seen in Figure~\ref{ModelAnalysis_IndividualStrengthsOfFeatureCategories} that $F_{1-3}$ (punctuation $n$-grams) are among the strongest feature categories. The degree of overlap between the distributions of $F_{1-3}$ is smaller in comparison to the other feature categories, while at the same time, the decision boundary can better separate them. When focusing on $F_4$ (TA sentence and clause starters), it can further be seen that this feature category is also important for these three corpora, while for $\CorpusYelp$ it plays only a minor role. With regard to $\CorpusYelp$, $F_1$ is primarily relevant, while the remaining feature categories contribute similarly to the overall prediction.  Moreover, for the mixed-topic corpora $\CorpusReddit$ and $\CorpusAmazon$, it can be seen that $F_6$ (TA token $1$-Grams) also represents a strong feature category, based on the observation that the distributions (in both $\CorpusReddit$ and $\CorpusAmazon$) only intersect in a small region around the decision boundary. 
\begin{table*} 
	\centering\small
	\begin{adjustbox}{max width=\textwidth}	
	\begin{tabular}{llllllllllll}
		\toprule
		%==================================================================================================
		\bfseries\boldmath Corpus $\Corpus$ & \boldmath$(F_{1}, \Threshold_{F_{1}})$ &  \boldmath$(F_{2}, \Threshold_{F_{2}})$ &  \boldmath$(F_{3}, \Threshold_{F_{3}})$ &  \boldmath$(F_{4}, \Threshold_{F_{4}})$ &  \boldmath$(F_{5}, \Threshold_{F_{5}})$ &  \boldmath$(F_{6}, \Threshold_{F_{6}})$ &  \boldmath$(F_{7}, \Threshold_{F_{7}})$ &  \boldmath$(F_{8}, \Threshold_{F_{8}})$ &  \boldmath$(F_{9}, \Threshold_{F_{9}})$ &  \boldmath$(F_{10}, \Threshold_{F_{10}})$ &  \boldmath$(F_{11}, \Threshold_{F_{11}})$ \\\midrule
		%==================================================================================================
		$\CorpusStackEx$ & $(F_1, 0.288)$ & $(F_2, 0.686)$ & $(F_3, 1.199)$ & $(F_4, 1.147)$ &  &  &  & $(F_8, 1.930)$ &  &  &  \\
		$\CorpusYelp$    & $(F_1, 0.504)$ &  & $(F_3, 1.663)$ & $(F_4, 1.475)$ & $(F_5, 2.000)$ & $(F_6, 1.058)$ & $(F_7, 1.859)$ & $(F_8, 1.986)$ &  & $(F_{10}, 1.871)$  &  \\
		$\CorpusReddit$  & $(F_1, 0.343)$ & $(F_2, 0.757)$ &  & $(F_4, 1.181)$ &  & $(F_6, 0.641)$ &  & $(F_8, 1.956)$ & $(F_9, 1.996)$ & $(F_{10}, 1.671)$ & $(F_{11}, 1.869)$ \\
		$\CorpusAmazon$  & $(F_1, 0.349)$ & $(F_2, 0.801)$ &  & $(F_4, 1.108)$ &  & $(F_6, 0.680)$ & $(F_7, 1.622)$ &  &  &  & $(F_{11}, 1.862)$ \\
		%==================================================================================================
		\bottomrule	
	\end{tabular}
	\end{adjustbox}
	\caption{Model analysis: Each row (starting at column two) represents a model $\bm{\Model}$ learned on the respective training corpus. \label{tab:OptimalFeatureCategories}}
\end{table*}

\subsubsection{Feature Analysis} \label{TAVeer_FeatureAnalysis}
\begin{figure}  % trim=[left bottom right top] 
	\centering
	\includegraphics[width=0.56\linewidth,trim=0.32cm 0.2cm 1.5cm 1.2cm,clip]{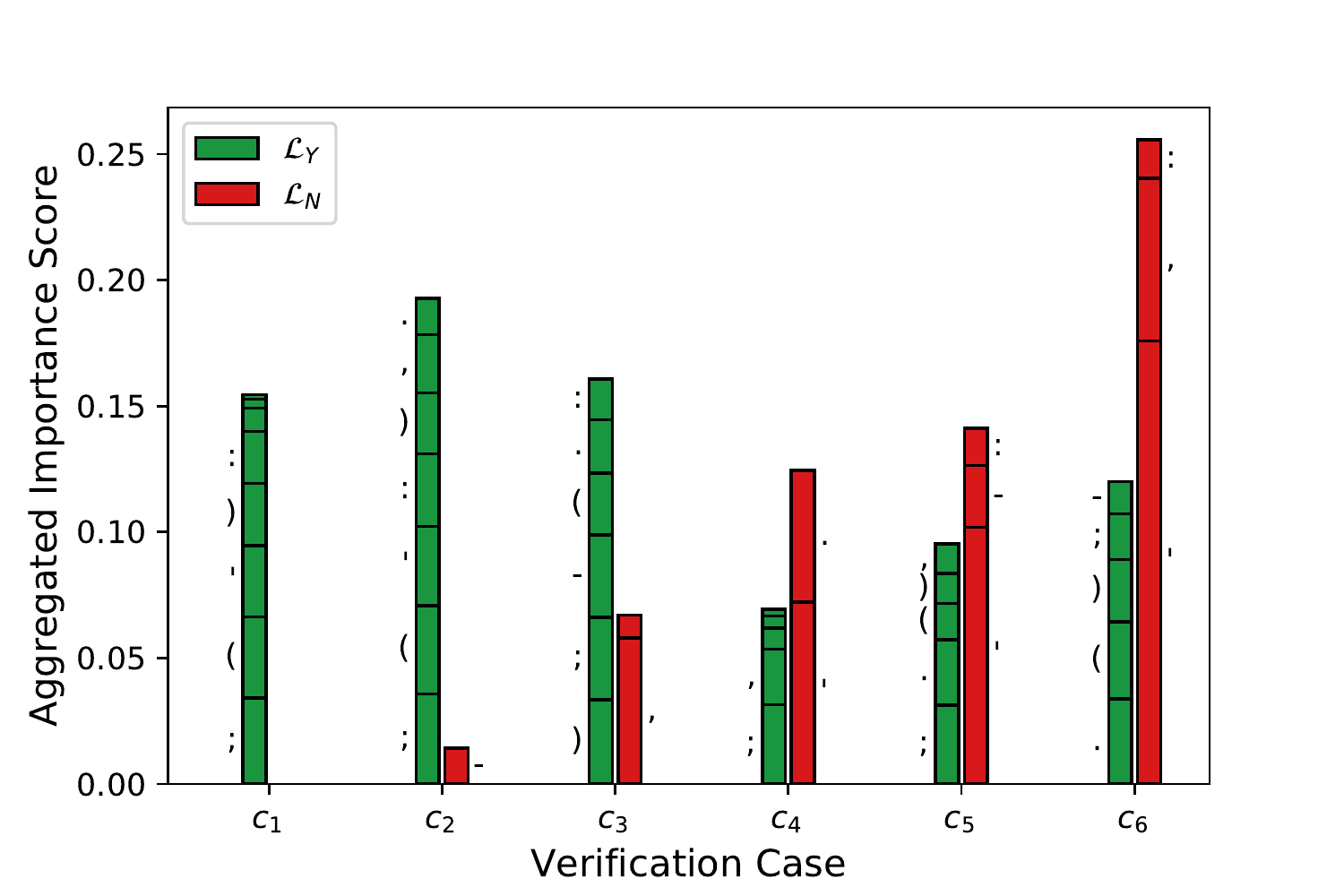} 
	\caption{Feature analysis regarding six verification cases $\bm{c_{1-6}}$ in the $\bm{\CorpusStackEx}$ test corpus (for the feature category $\bm{F_1}$). \label{fig:FeatureAnalysis}}
\end{figure}

So far, we investigated the question which particular feature categories were more relevant regarding the \numberOfTestCorpora test corpora. 
In the following, we take a closer look at the classification results of \taveer with respect to six verification cases $(c_1, c_2, \ldots, c_6)$ originating from the test corpus $\CorpusStackEx$. Here, the cases $c_{1-3}$ and $c_{4-6}$ were correctly classified as \classY (true positives) and \classN (true negatives), respectively. To perform the analysis, we use our interpretation method introduced in Section~\ref{TAVeer_InterpretationScheme}. 
%--------------------------------------------------------------------------------------------------------------- 
Recall that this method consumes as input a verification case $\Problem = (\Dunk, \DA)$, an atomic ensemble $(F, \Threshold_{F})$, and a distance function dist$(\cdot)$. 
The output are two disjunct lists \listY and \listN comprising tuples of the form $(v, \featureImp)$, where $v$ represents a feature and $\featureImp$ its corresponding importance score. 
%--------------------------------------------------------------------------------------------------------------- 
Given these tuples, we can see which features (and to what extent) contributed either to the \classY- or \classN-prediction of \taveer.  
In our model analysis we found that $F_1$ was one of the strongest feature categories across all test corpora. 
Therefore, we focus on this feature category in the following.  
%---------------------------------------------------------------------------------------------------------------
After applying Algorithm~\ref{AVeerInterpretationScheme} to the six verification cases, we obtain for each $\Problem_i$ a pair of two lists \listY and \listN. 
The impact of these features are visualized in Figure~\ref{fig:FeatureAnalysis}. 
%---------------------------------------------------------------------------------------------------------------
%--------------------------------------------------------------------------------------------------------------- 
From the illustration in Figure~\ref{fig:FeatureAnalysis} a number of observations can be made. 
Regarding $\Problem_1$, for example, it can be seen that \listN-features are missing, which explains why $\Problem_1$ achieved the highest similarity value of all verification cases. While inspecting the occurrences of the \listY-features \textcolor{Black}{\textbf{\texttt{;('):-}}} in the documents within $c_1$, we noticed that their frequency distributions were almost equal. 

For $\Problem_2$, it can be noticed that almost each \listY-feature is more important than the single \listN-feature. In other words, the hyphen \textcolor{Black}{\textbf{\texttt{-}}} is the only feature that led to a (small) discrepancy between the frequency distributions of all punctuation marks in $\Dunk$ and $\DA$ within $\Problem_2$. 
For $c_3$, we can see that the six punctuation marks \textcolor{Black}{\textbf{\texttt{();-.:}}} are sufficient for its correct \classY-prediction. In contrast, only the two punctuation marks \textcolor{Black}{\textbf{\texttt{.'}}} are required to correctly classify $c_4$ as \classN. With regard to $c_{4-6}$ it is further shown that the \listN-feature \textcolor{Black}{\textbf{\texttt{'}}} is the most important feature, which alone is sufficient to distinguish the authorships in all three cases. The importance score of this feature points to a greater usage of contractions, which are only (or mostly) present in one document. Overall, we can conclude from these observations that correct predictions can be achieved with just a few features. 

%% file: texfiles/Conclusions.tex
\section{Conclusion and Future Work} \label{Conclusions}
%--------------------------------------------------------------------------------------------------------------------------------------
% Welches wissenschaftliche Problem haben wir im Paper untersucht?
%--------------------------------------------------------------------------------------------------------------------------------------
We discussed an important problem in the field of \av, which occurs when an \AV method has no control over the features it captures. 
In the worst case, the prediction of the method may be based on topic-related words rather than on stylistic features, so that the \AV method will miss its true purpose. 
%--------------------------------------------------------------------------------------------------------------------------------------
% Wie gehen wir das Problem an?
%--------------------------------------------------------------------------------------------------------------------------------------
To address this problem, we have made three contributions. 
First, we proposed a number of feature categories that comprise a wide spectrum of topic-agnostic (TA) features. 
Second, we proposed an effective distance-based \AV method called \taveer, which considers solely these feature categories for its classification predictions. 
Third, we proposed an interpretation scheme that allows to understand which features contributed to \taveer's prediction. 
 
%--------------------------------------------------------------------------------------------------------------------------------------
% Wie haben wir das Verfahren evaluiert? 
%-------------------------------------------------------------------------------------------------------------------------------------- 
To assess our approach, we performed a comprehensive evaluation with \numberOfExistingBaselines existing \AV approaches applied to \numberOfTestCorpora corpora with related-, cross- and mixed topics. In this regard, we have demonstrated that our method outperforms all baseline methods with respect to one cross-topic and one mixed-topic corpus, while on the other two corpora, \taveer performs close to the strongest baseline \coav. 
%--------------------------------------------------------------------------------------------------------------------------------------
In a detailed analysis with respect to the models learned by \taveer, we have shown that punctuation $n$-grams (for $n\in \{1,2,3\}$), TA sentence and clause starters and TA token unigrams were the strongest feature categories across all \numberOfTestCorpora test corpora. Furthermore, we have shown in a fine-grained feature analysis that a small number of features is sufficient to correctly classify challenging cross-topic verification cases (two and six features for a \classN- and \classY-case, respectively). 

%--------------------------------------------------------------------------------------------------------------------------------------
% Limitations & Future work.
%--------------------------------------------------------------------------------------------------------------------------------------
Nevertheless, our \AV method leaves room for further improvements. 
Currently, \taveer does not take into account misspelled words, which can lead to a loss of potentially relevant features, especially in connection with informal texts. 
We therefore leave for future work the investigation of effective possibilities to semantically match misspelled words with respect to their common entity. 
One idea, for example, is to use back-translation services that can handle difficult spelling mistakes, which cannot be corrected by standard spell checkers.  
Furthermore, we plan to find a better alternative to visualize the features derived from the interpretation scheme more clearly (ideally in the document itself). 
However, this requires a careful consideration how to map the importance scores of the features (across all feature categories) to respective colors without to cause overplotting. 
%--------------------------------------------------------------------------------------------------------------------------------------
A further direction for future work is to investigate alternative feature categories not yet been considered in this paper. 
In this context, one idea is to experiment with interjections (\eg \texttt{"lol"} or \texttt{"aha"}) or topic-agnostic abbreviations (for example, \texttt{"e.g."} or \texttt{"etc."}), which represent important idiosyncratic stylistic markers. 
%--------------------------------------------------------------------------------------------------------------------------------------
Another question we want to address in future work is how \taveer behaves under cross-domain conditions. These are of particular relevance as they often occur in real forensic cases (\eg how a model trained on forum posts or cooking recipes performs on suicide letters, for which no training data is available).

%% file: texfiles/Acknowledgments.tex
\section{Acknowledgments}
This research work has been funded by the German Federal Ministry of Education and Research and the Hessen State Ministry for Higher Education, Research and the Arts within their joint support of the National Research Center for Applied Cybersecurity ATHENE. We would like to thank Christian Winter and Inna Vogel for their valuable reviews that helped to improve the quality of this paper.